\pdfoutput=1
\documentclass{article}

\usepackage[final]{CPAL-2025-template/cpal_2025}

\usepackage{hyperref}
\usepackage{url}
\usepackage{xspace}

\usepackage{booktabs}       % professional-quality tables
\usepackage{amsfonts}       % blackboard math symbols
\usepackage{nicefrac}       % compact symbols for 1/2, etc.
\usepackage{microtype}      % microtypography
\usepackage{xcolor}         % colors
\usepackage{xspace}
\usepackage{natbib}

\usepackage{graphicx}
\usepackage{pifont}
\usepackage{tcolorbox}
\usepackage{tabularx}
\usepackage{makecell}
\usepackage{multirow}
\usepackage{multicol}
\usepackage{amssymb}
\usepackage{bbm}
\usepackage{amsmath}
\usepackage{wrapfig}
\usepackage{cleveref}
\usepackage{enumitem}
\usepackage{titletoc}

\usepackage{mathtools}
\usepackage{amsthm}
\newtheorem{lemma}{Lemma}
\usepackage{array}
\usepackage{caption}
\usepackage{algorithm}
\usepackage{algorithmic}
\usepackage{graphicx}
\usepackage{booktabs}
\usepackage{xcolor}
\usepackage{multirow}
\usepackage{amsfonts}
\usepackage{xcolor}
\usepackage{colortbl}
\usepackage{colortbl}
\usepackage{makecell}
\usepackage{float}
\usepackage{pifont}
\usepackage{placeins}
\usepackage{wrapfig}
\usepackage{tabularx}
\usepackage{subcaption}
\usepackage[english]{babel}      % 开启英文断词规则
\usepackage{array}               % 支持自定义列格式
\usepackage[table]{xcolor}
\newcolumntype{R}[1]{>{\raggedright\arraybackslash}p{#1}} % 定义带自动换行的左对齐列
\definecolor{lightergreen}{RGB}{228, 108, 10}

\definecolor{darkgreen}{rgb}{0.0, 0.5, 0.0}
\definecolor{text_red}{RGB}{220,20,60}

\hypersetup{
  colorlinks,
  linkcolor={red!90!black},
  citecolor={orange!90!black},
  urlcolor={blue!70!black}
}

\title{ERC-SVD: Error-Controlled SVD for Large Language Model Compression}

\author{%
    Haolei Bai\textsuperscript{1,2},
    ~Siyong Jian\textsuperscript{1,4},
    ~Tuo Liang\textsuperscript{3},
    ~Yu Yin\textsuperscript{3},
    ~Huan Wang\textsuperscript{1}\thanks{Corresponding author: \texttt{wanghuan@westlake.edu.cn}}\\
  \textsuperscript{1}Westlake University, \textsuperscript{2}Nanyang Technological University, \\ \textsuperscript{3}Case Western Reserve University, \textsuperscript{4}Nanjing University\\
}

\begin{document}

\maketitle

\begin{abstract}
Large language models (LLMs) have demonstrated impressive capabilities in a wide range of downstream natural language processing tasks. 
Nevertheless, their considerable sizes and memory demands hinder practical deployment, underscoring the importance of developing efficient compression strategies. 
Singular value decomposition (SVD) decomposes a matrix into orthogonal components, enabling efficient low-rank approximation. This is particularly suitable for LLM compression, where weight matrices often exhibit significant redundancy.
However, current SVD-based methods neglect the residual matrix from truncation, resulting in significant truncation loss. 
Additionally, compressing all layers of the model results in severe error propagation. 
To overcome these limitations, we propose \textbf{ERC-SVD}, a new post-training SVD-based LLM compression method from an error-controlled perspective. 
Specifically, we leverage the residual matrix generated during the truncation process to reduce truncation loss. 
Moreover, under a fixed overall compression ratio, we selectively compress the last few layers of the model, which mitigates error propagation and improves compressed model performance.
Comprehensive evaluations on diverse LLM families and multiple benchmark datasets indicate that ERC-SVD consistently achieves superior performance over existing counterpart methods, demonstrating its practical effectiveness.

{\renewcommand\twocolumn[1][]{#1}
\begin{center}
    \centering
    \renewcommand{\arraystretch}{0.05} %
    \vspace{-2mm}
    \begin{tabular}{c}
        \hspace{-1.6cm}
        % , height = 0.35\linewidth
        \includegraphics[width = 0.39\linewidth]{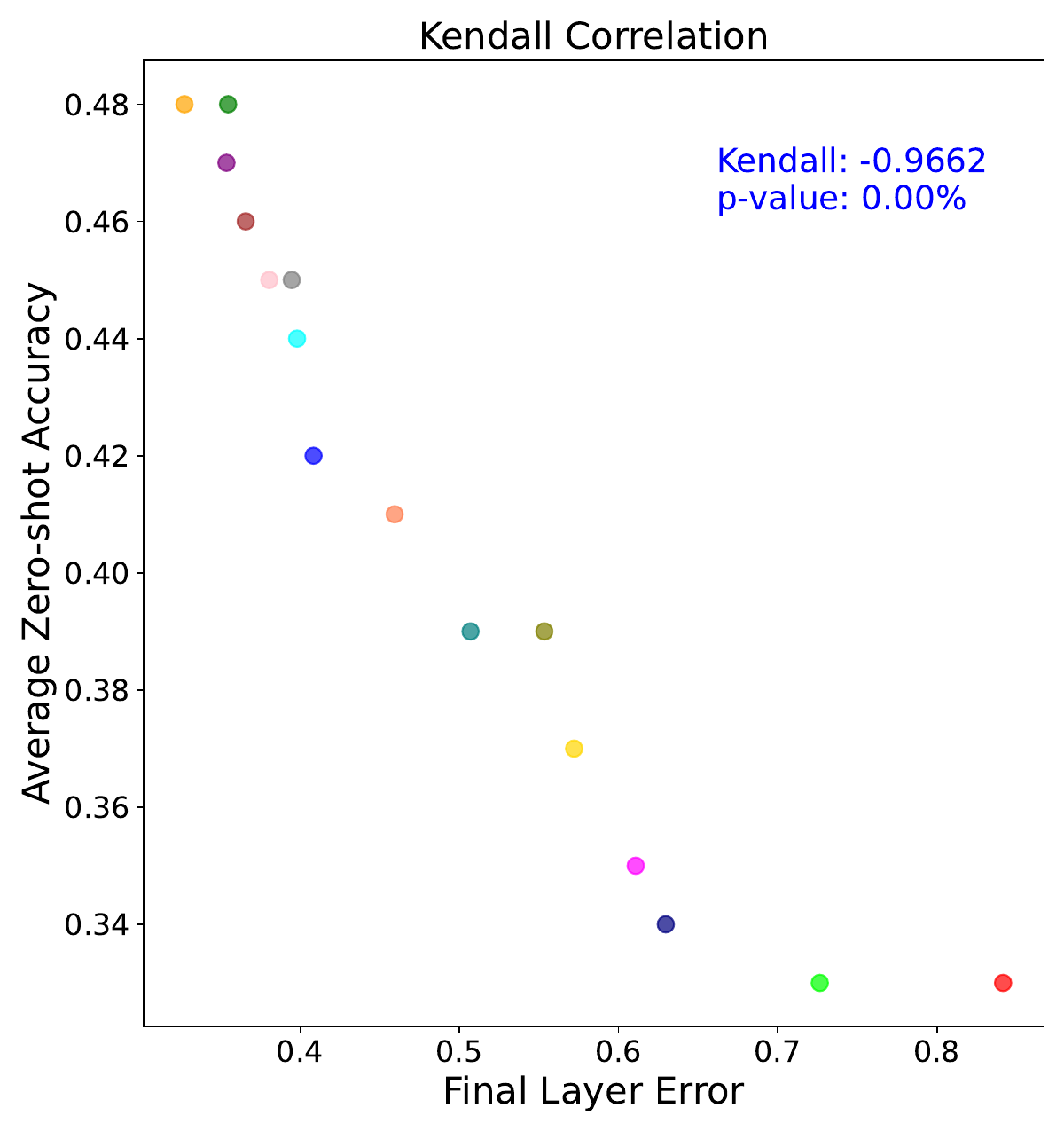} 
        \hspace{0cm}
        % ,height = 0.3\linewidth
        \includegraphics[width = 0.39\linewidth]{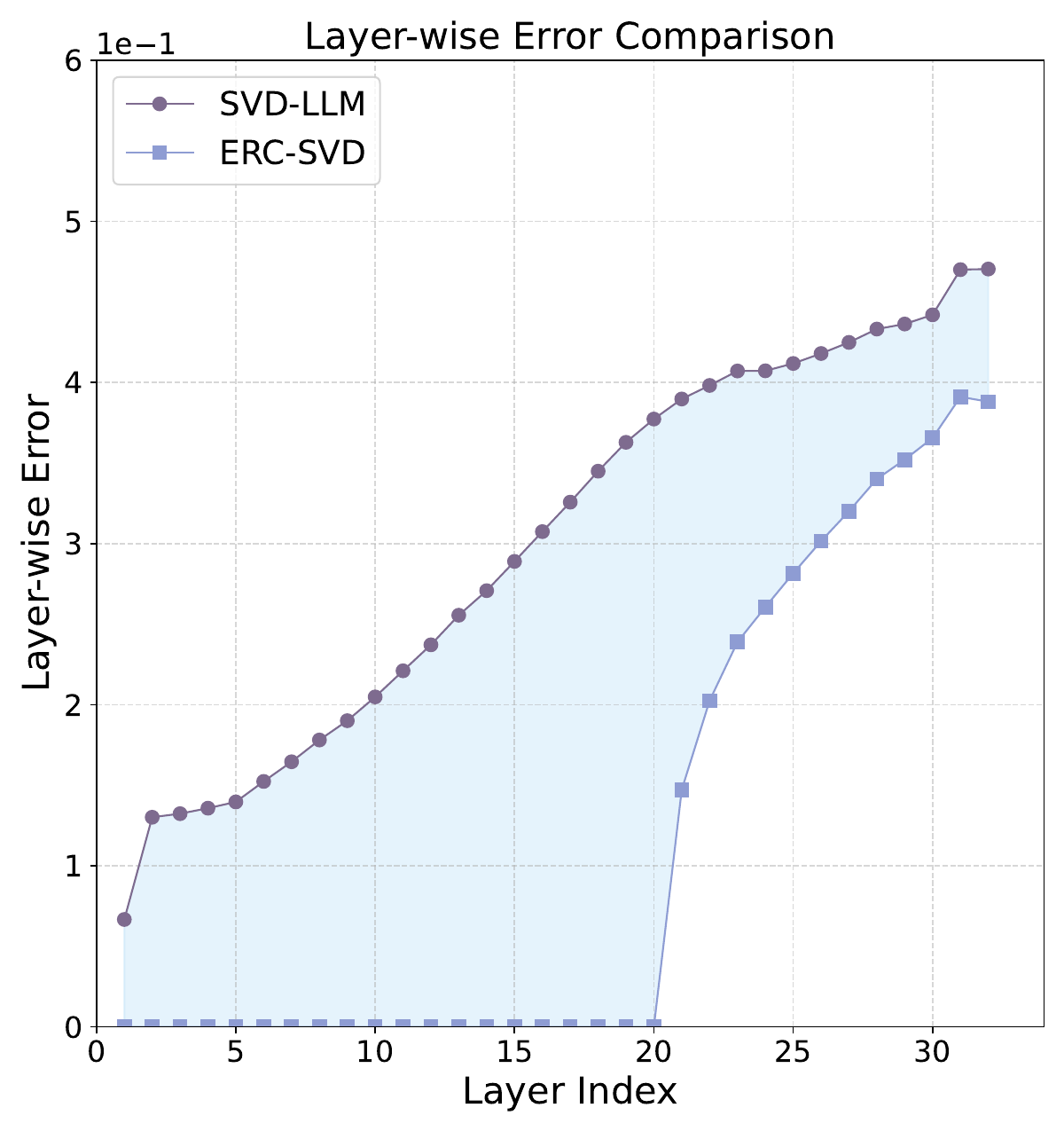}
        \hspace{0cm}
        \includegraphics[width = 0.39\linewidth]{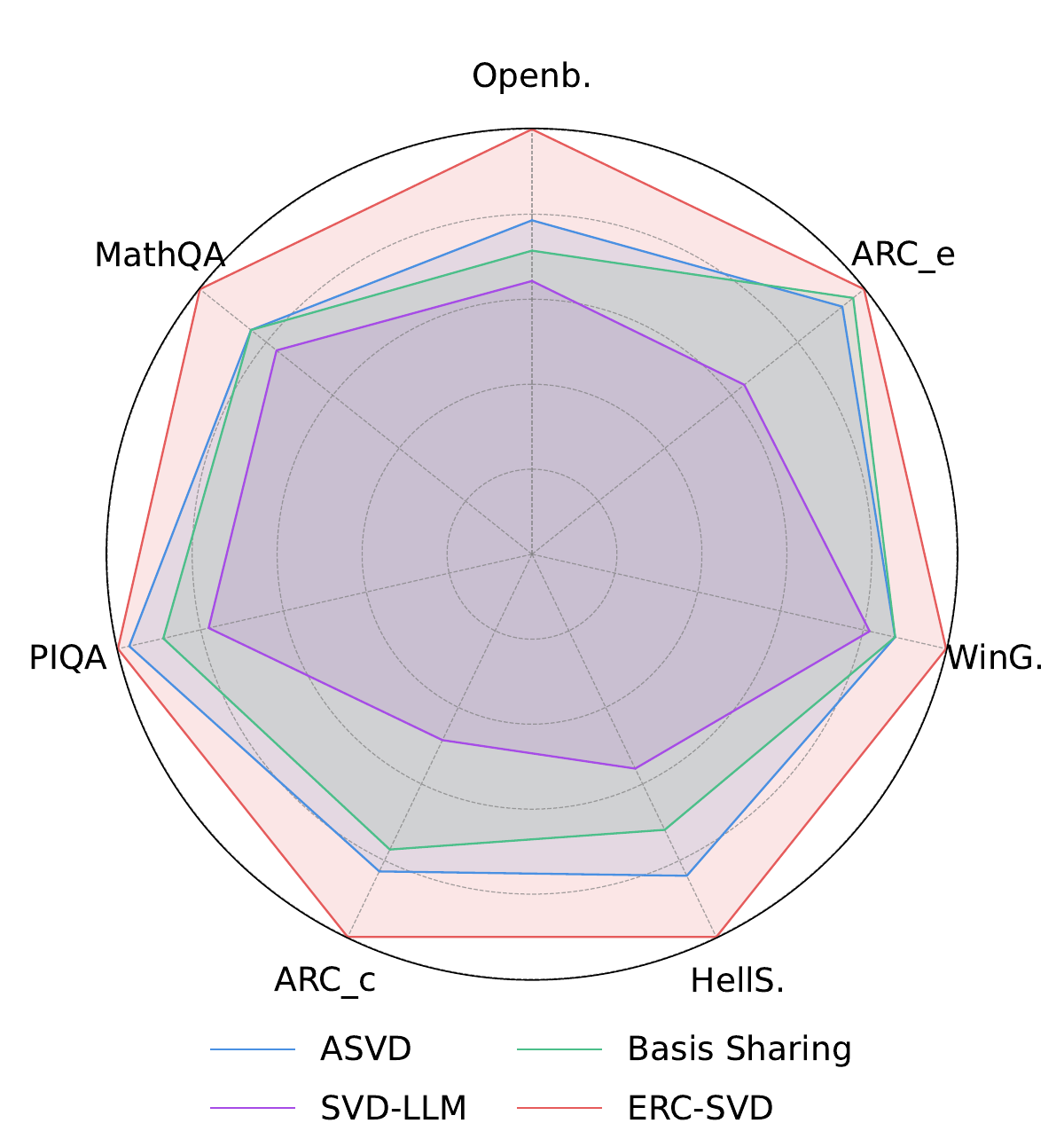}
        \end{tabular}
        \captionof{figure}{Left: The Kendall correlation between the final-layer error and the average zero-shot accuracy. Middle: Layer-wise error comparison between ERC-SVD and SVD-LLM of LLaMA-2-7B under 20\% compression ratio. Right: The accuracy comparison of LLaMA-2-7B compressed by different methods under 20\% compression ratio on seven reasoning and understanding tasks.}
    \label{fig:teaser}
\end{center}
}
\end{abstract}
\section{Introduction}
Large language models (LLMs) have emerged as powerful tools, delivering state-of-the-art performance across a wide range of tasks such as text generation, translation, and reasoning.
The scaling law~\cite{kaplan2020scaling} has driven a trend toward increasingly large models, exemplified by models such as GPT~\cite{brown2020language}, PaLM~\cite{chowdhery2023palm}, LLaMA~\cite{touvron2023llama}, Deepseek~\cite{liu2024deepseek}, and Qwen~\cite{yang2025qwen3}, which often contain tens to hundreds of billions of parameters. 
Despite their powerful capabilities, the enormous scale of LLMs poses serious challenges for efficient deployment due to high computational demands~\cite{sheng2023flexgen,zhou2024survey,wang2024model}. 
This resource burden not only limits deployment on edge devices and consumer-level hardware but also increases the cost and carbon footprint of serving LLMs in production~\cite{strubell2020energy,patterson2022carbon}.

As the scale of LLMs continues to grow, compression techniques, including weight quantization~\cite{frantar2022gptq,lin2024awq,huang2024billm,li2024arb}, network pruning~\cite{ma2023llm,frantar2023sparsegpt,sunsimple,gao2024disp,su2026rose}, knowledge distillation~\cite{gu2023minillm,yang2024survey,xu2024survey,zhang2024dual}, and low-rank decomposition~\cite{hsulanguage,kaushal2023lord,yuan2023asvd,wang2024svd,li2025adasvd}, have become increasingly important for the practical deployment of LLMs in resource-constrained environments. 
The syntactic and semantic correlations acquired during training often induce redundancy in LLM weight matrices, giving rise to a low-rank structure~\cite{saha2024compressing}. 
As a result, singular value decomposition (SVD) provides a principled and effective approach for compressing these matrices with minimal loss. 
In particular, post-training approaches are gaining traction, as they can significantly reduce memory and compute requirements without the need for expensive retraining, making them especially suitable for scaling up foundation models. 
Recent studies on post-training SVD-based LLM compression, including ASVD~\cite{yuan2023asvd}, SVD-LLM~\cite{wang2024svd}, Basis Sharing~\cite{wang2024basis}, and AdaSVD~\cite{li2025adasvd}, have made significant progress in reducing model size while preserving performance, demonstrating the effectiveness of low-rank approximation techniques in compressing LLMs. 
However, existing methods suffer from two major limitations. First, existing methods ignore the importance of the residual matrix generated during SVD truncation, leading to significant truncation loss. Second, compressing all model layers often results in high layer-wise error and severe error propagation.

In this work, we propose \textbf{ERC-SVD}, a new post-training SVD-based compression method for LLMs. 
Building upon the key observations outlined above, ERC-SVD introduces two core technical innovations.
\ding{172} \textbf{Residual compensation for SVD truncation:} The residual matrix produced during SVD truncation can be effectively utilised to reduce the truncation loss. 
Specifically, we perform SVD truncation in two stages: we first truncate the original weight matrix $\boldsymbol{W}$ to obtain its intermediate low-rank approximation $\boldsymbol{W}_{r_i}$. After that, we compute the residual matrix $\boldsymbol{R}$ between $\boldsymbol{W}$ and $\boldsymbol{W}_{r_i}$. 
Second, we apply SVD truncation to $\boldsymbol{R}$, yielding $\boldsymbol{R}_{r_r}$. 
Finally, we construct the compressed weight matrix $\boldsymbol{\hat{W}}_{r}=\boldsymbol{W}_{r_i}+\boldsymbol{R}_{r_r}$, where the rank satisfies $r_i + r_r = r$. Detailed description and a mathematical proof are provided in Section~\ref{sec_rc}. \ding{173} \textbf{Partial-layer compression for SVD:} LLMs consist of a sequence of consecutive layers, where the output of each layer serves as the input to the next. 
Therefore, any error introduced in earlier layers can propagate and accumulate through subsequent layers, leading to severe performance degradation. 
To mitigate this, we propose compressing only the last few layers under a fixed overall compression ratio while keeping the earlier layers intact. This strategy ensures that the earlier layers remain error-free, reducing the impact of error propagation.

Our key contributions can be summarised as follows:
\begin{itemize}
    \item We introduce \textbf{residual compensation for SVD truncation}, a theoretically grounded compensation strategy. By leveraging the residual matrix generated during the SVD truncation, our strategy significantly reduces the overall truncation loss.
    \item We propose \textbf{partial-layer compression for SVD}, which compresses only the last few layers of the model under a fixed overall compression ratio. This strategy effectively reduces layer-wise error and mitigates error propagation.
    \item Extensive experiments on multiple LLMs (LLaMA, OPT, Mistral, Vicuna, and Qwen) and benchmark datasets (both language modelling and zero-shot reasoning) demonstrate that ERC-SVD outperforms existing methods across a wide range of compression ratios.
\end{itemize}
\section{Related Work}
\subsection{Techniques for Large Language Model Compression}
The growing size of large language models (LLMs) has raised increasing concerns over their computational and memory demands. 
To address these challenges, a variety of model compression techniques have been proposed. 
Conventional approaches often require computationally expensive retraining, which is generally impractical due to the substantial computational cost associated with the massive size of LLMs. 
Consequently, recent efforts have shifted toward more resource-friendly post-training compression techniques~\cite{frantar2022gptq,frantar2023sparsegpt,zhu2024survey}. 
Typically employed approaches include network pruning and weight quantization. And pruning techniques can be classified into unstructured and structured methods.
Unstructured pruning removes individual weights based on importance scores. 
For example, SparseGPT~\cite{frantar2023sparsegpt} performs one-shot pruning using second-order approximations without retraining.  
However, since unstructured pruning retains the original matrix shape, it offers limited inference acceleration and requires specialized hardware. 
In contrast, structured pruning eliminates entire blocks or channels, enabling compatibility with conventional hardware platforms. 
LLM-Pruner~\cite{ma2023llm} groups dependent linear projections into coupled structures, assigns each group a loss‑aware importance score, prunes the least important groups, and applies LoRA fine‑tuning to restore performance. Additionally, ZipLM~\cite{kurtic2023ziplm} prioritizes pruning components that yield the worst trade-off between latency and loss, but this often causes notable performance degradation. 
Quantization provides another mainstream solution. 
GPTQ~\cite{frantar2022gptq} gradually quantizes and updates each weight using the Hessian matrix to minimize the quantization error. 
AWQ~\cite{lin2024awq} preserves important weight channels by selecting reparameterization coefficients via grid search. Furthermore, BiLLM~\cite{huang2024billm} and ARB-LLM~\cite{li2024arb} push quantization to the 1-bit level, while still delivering impressive downstream task accuracy. 
However, these techniques still cause significant performance degradation, especially at low bit-widths, due to the lack of weight and activation adaptation. 

\subsection{SVD-based Techniques for Compressing LLMs}
Singular value decomposition (SVD) is a commonly employed technique for reducing matrix dimensionality by representing the original matrix as the product of two low-rank factor matrices. 
Recent research has demonstrated the potential of SVD-based LLM compression methods, yet comprehensive exploration remains limited. 
Hsu et al.~\cite{hsulanguage} propose FWSVD, incorporating Fisher information to weight the importance of parameters. 
However, it relies on computationally intensive training and was originally applied only to small language models (e.g., BERT~\cite{devlin2019bert}, ALBERT~\cite{lan2019albert}). 
LoRD~\cite{kaushal2023lord} first applies SVD to LLMs by grouping layers to improve efficiency. 
However, LoRD overlooks the importance of input activations.
To address this limitation, Yuan et al.~\cite{yuan2023asvd} introduce ASVD, which mitigates the impact of activation outliers by reshaping the weight matrix based on activation distribution, thereby enhancing the precision of the low-rank decomposition. 
CALDERA~\cite{saha2024compressing} exploits the inherent low-rank structure of weight matrices by approximating them through a low-rank, low-precision decomposition.
Additionally, SVD-LLM~\cite{wang2024svd} establishes a direct mapping between singular values and truncation loss, which means that truncating the smallest singular values leads to minimal truncation loss.
Basis Sharing~\cite{wang2024basis} achieves more effective LLM compression through sharing parameters across different layers. 
More recently, AdaSVD~\cite{li2025adasvd} proposes an adaptive compensation method, iteratively updating matrices during truncation. 

Despite these advances, all existing methods ignore the residual matrix produced during SVD truncation, which can significantly compensate for the SVD truncation loss. 
Moreover, compressing only the last few layers of the model under a fixed target compression ratio can provide better performance for compressed models. 
However, existing methods compress all layers, either apply a uniform compression ratio across all layers or assign variable ratios based on layer-wise importance, which often results in sub-optimal performance.

\section{ERC-SVD} \label{ER-SVD_method}

\begin{figure}[ht]
    \centering
    \includegraphics[width=\textwidth]{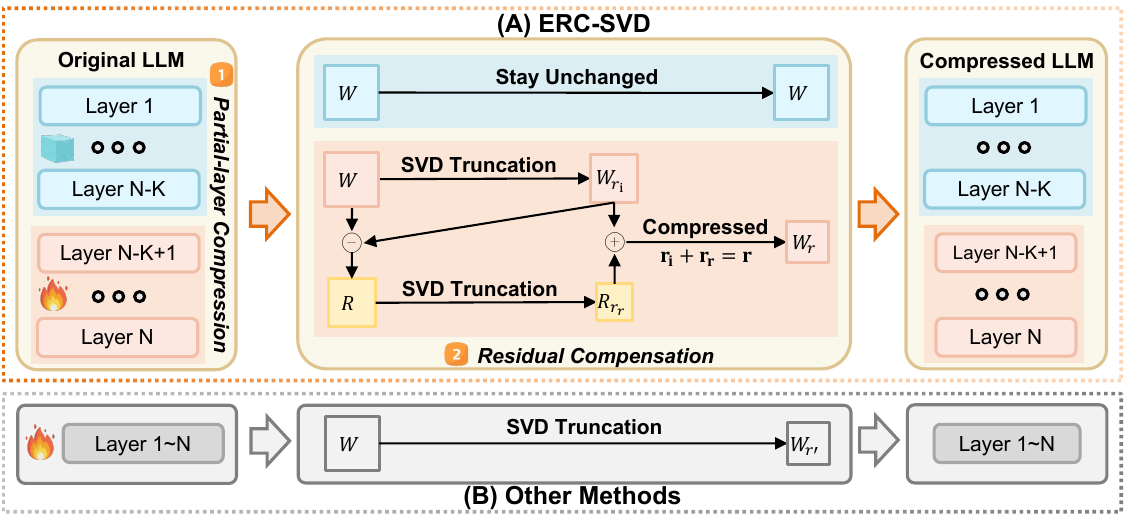}
    \caption{The overall framework of ERC-SVD, and comparison with other methods. The last $k$ layers are selected through \textbf{partial-layer compression} and compressed using \textbf{residual compensation} with calibration data. \includegraphics[height=0.8em]{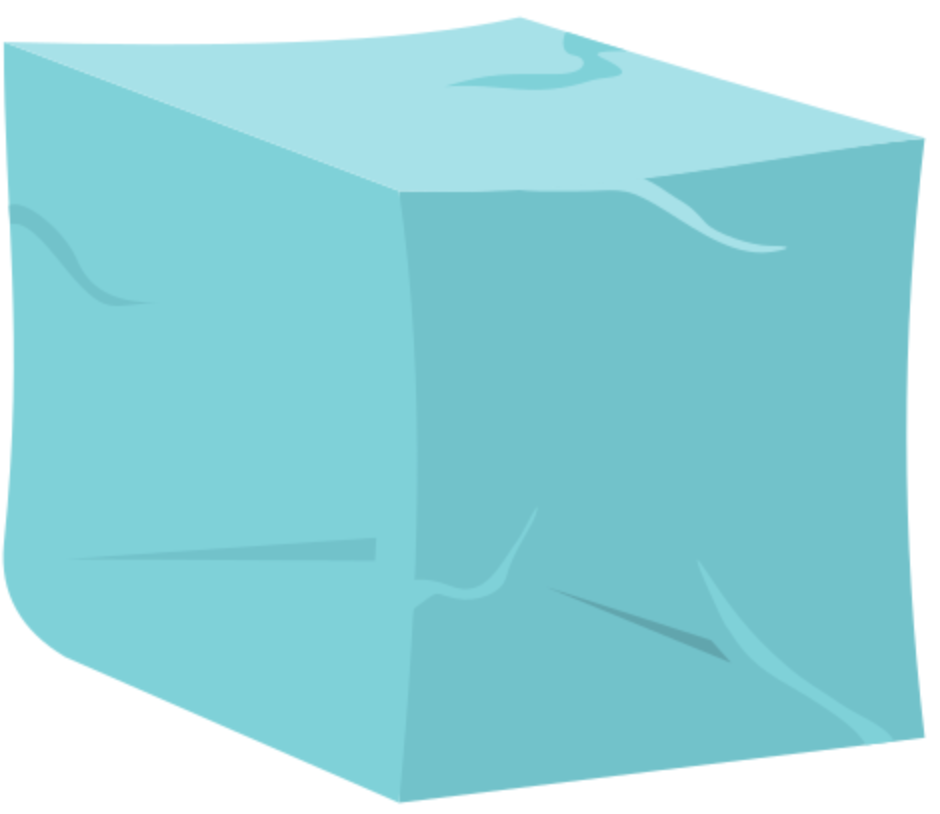} denotes these layers remain intact, while \includegraphics[height=1em]{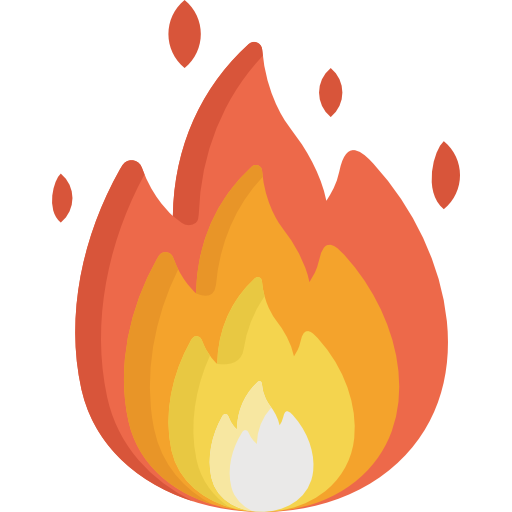} denotes these layers are replaced by low-rank approximations. The overall compression ratio is $R_o$, for ERC-SVD, the first $(N-k)$ layers stay unchanged, and the layer compression ratio $R_l$ for last $k$ layers is $(N \cdot R_o)/k$.} \label{ER-SVD_framework}
\end{figure}

The framework of ERC-SVD is illustrated in Figure~\ref{ER-SVD_framework}. 
We first perform SVD to compress the last few layers of the model, while ensuring that the overall compression ratio is satisfied, and compute the corresponding residual matrix. 
A second SVD is then applied to the residual matrix to obtain its low-rank approximation.
The two truncated matrices are subsequently combined to construct the final compressed weight matrices.
Compared with other methods~\cite{yuan2023asvd,wang2024svd,wang2024basis,li2025adasvd}, we utilize the residual matrix during the truncation process to compensate for the truncation loss.
In Section~\ref{sec_rc}, we first provide a detailed description of how \textbf{residual compensation} effectively works. 
Subsequently, in Section~\ref{sec_plc}, we explain the benefits of \textbf{partial-layer compression} in mitigating error propagation. The pseudocode of ERC-SVD is shown in Algorithm~\ref{alg_ER-SVD}, with the pseudocode for residual compensation and partial-layer compression provided in Algorithm~\ref{alg_rc} and Algorithm~\ref{alg_plc}, respectively.

\subsection{Residual Compensation for SVD Truncation}\label{sec_rc}
\paragraph{Preliminaries for SVD.} SVD-based compression methods apply SVD to the original weight matrix $\boldsymbol{W}\in\mathbb{R}^{m\times n}$, discarding the smallest singular values to obtain its low-rank approximation $\boldsymbol{W}_r$:
\begin{align}
   \boldsymbol{W} = \boldsymbol{U}\boldsymbol{\Sigma} \boldsymbol{V}^T \approx \boldsymbol{U}_r\boldsymbol{\Sigma}_r\boldsymbol{V}_r^T = \boldsymbol{W}_r\;,r < \min(m,n),
\end{align}
where $\boldsymbol{U}_r\in\mathbb{R}^{m\times r}$ and $\boldsymbol{V}_r^T\in\mathbb{R}^{r\times n}$ are composed of the top-$r$ left and right singular vectors, respectively, and $\boldsymbol{\Sigma}_r\in\mathbb{R}^{r\times r}$ is a diagonal matrix containing the corresponding singular values. 
Following prior post-training SVD-based works~\cite{yuan2023asvd,wang2024svd}, the optimization objective for SVD truncation in LLMs can be formulated as:
\begin{gather}
   \hat{\boldsymbol{W}}_r = \arg\min_{\boldsymbol{W}_r}\big\|\boldsymbol{WX}-\boldsymbol{W}_r \boldsymbol{X}\big\|_F \;, \\
   \mathcal{L} = \big\|\boldsymbol{WX}-\boldsymbol{W}_r \boldsymbol{X}\big\|_F = \big\|(\boldsymbol{W}-\boldsymbol{W}_r)\boldsymbol{X}\big\|_F \;, \label{eq4}
\end{gather}
where $\boldsymbol{X}$ denotes the activation of $\boldsymbol{W}$ given an input, and $\mathcal{L}$ is the truncation loss measured by the Frobenius norm. 
Although previous works~\cite{hsulanguage,yuan2023asvd,wang2024svd} have made significant progress in minimizing $\mathcal{L}$, they consistently overlook the residual matrix generated. Based on Equation~\ref{eq4}, minimizing the truncation loss reduces to minimizing the discrepancy between the original weight matrix $\boldsymbol{W}$ and its low-rank approximation $\boldsymbol{W}_r$. Accordingly, we reformulate the optimization objective as:
\begin{align}
      \hat{\boldsymbol{W}}_r = \arg\min_{\boldsymbol{W}_r}\big\|\boldsymbol{W}-\boldsymbol{W}_r\big\|.
\end{align}

\begin{algorithm}[ht]
    \caption{Pseudocode of ERC-SVD}
    \label{alg_ER-SVD}
    \renewcommand{\algorithmicrequire}{\textbf{Input:}}
    \renewcommand{\algorithmicensure}{\textbf{Output:}}
    
    \begin{algorithmic}[1]
        \REQUIRE Original LLM: $M$, weight matrix: $\boldsymbol{W}\in\mathbb{R}^{m\times n}$, number of model layers: $N$, residual compensation rank $r_r$, step: $s$  %%input
        \ENSURE Compressed LLM $M'$ by ERC-SVD    %%output
        
        \STATE $CD \gets$ Randomly collect calibration samples from the dataset
        
        \STATE $\mathrm{Set}_{\boldsymbol{W}} \gets M$, $\mathrm{Set}_{\boldsymbol{W'}} \gets \emptyset$ \hfill $\triangleright$ Initialize the sets of weight matrices

        \STATE $k$, $R_l \gets$ \textsc{Partial-layer Compression}($M$, $N$, $R_o$, $s$)

        %  \STATE $r = m \cdot n \cdot(1-LCR^{\star})/(m+n) $, $r_i = r-r_r$
        
        \FOR{Layer $i$ in original LLM $M$}
            \IF {$i \in [1, N-k)$}
                \STATE $\mathrm{Set}_{\boldsymbol{W'}}(i) \gets \mathrm{Set}_{\boldsymbol{W}}(i)$ \hfill $\triangleright$ Current weight matrices stay the same
            \ELSE
                \STATE $\boldsymbol{W}_i \gets$ $\mathrm{Set}_{\boldsymbol{W}}(i)$ \hfill $\triangleright$ Initialize weight matrices to compress
                
                \STATE $\mathrm{Set}_{\boldsymbol{W'}}(i) \gets$ \textsc{Residual Compensation}($M$, $\boldsymbol{W}_i$, $R_l$, $r_r$)
            \ENDIF

            \STATE $\mathrm{Set}_{\boldsymbol{W'}}$ $\gets$ $\mathrm{Set}_{\boldsymbol{W'}}(i)$ $\cup$ $\mathrm{Set}_{\boldsymbol{W'}}$ \hfill $\triangleright$ Append weight matrices after operation
            
        \ENDFOR
        \STATE $M'$ $\gets$ \textsc{Update}($M$, $CD$, $\mathrm{Set}_{\boldsymbol{W'}}$)
        
        \RETURN $M'$
    \end{algorithmic}
\end{algorithm}

\paragraph{Explanation for Residual Compensation.} 
Given the original weight matrix $\boldsymbol{W}\in\mathbb{R}^{m\times n}$ and the layer compression ratio $R_l$, the compression ratio in this work denotes the fraction of parameters removed. We define the scale of the matrix as: $\alpha = (m \cdot n) / (m+n)$, where $m$ and $n$ represent the input and output dimensions of the matrix, hence, the target rank for each layer is $r = (1-R_l)\cdot\alpha$. We introduce a residual compensation factor $\beta$, a hyperparameter fixed at $0.05$ in all experiments. The target rank $r$ is decomposed into two components: a residual compensation rank $r_r$, defined as $r_r = \alpha \cdot \beta$, and an intermediate rank $r_i$, defined as $r_i = r - r_r$. 

The entire compression process comprises two SVD truncations.
First, we apply SVD to $\boldsymbol{WS}$, where $\boldsymbol{S}$ is the scaling weight matrix obtained from the input activation, and retain the top-$r_i$ singular values: $\boldsymbol{WS} \approx \boldsymbol{W}_{r_i}' = \boldsymbol{U}_{r_i}\boldsymbol{\Sigma}_{r_i}\boldsymbol{V}_{r_i}^{'T} $, and multiply $\boldsymbol{S^{-1}}$ to obtain an intermediate low-rank approximation of the original weight matrix: $ \boldsymbol{W}_{r_i} = \boldsymbol{W}_{r_i}'\boldsymbol{S^{-1}} = \boldsymbol{U}_{r_i}\boldsymbol{\Sigma}_{r_i}\boldsymbol{V}_{r_i}^T. $
The residual matrix $\boldsymbol{R}$ is computed as the difference between $\boldsymbol{W}$ and $\boldsymbol{W}_{r_i}$, where 
$\boldsymbol{R} = \boldsymbol{W} - \boldsymbol{W}_{r_i}.$
We then perform a second SVD on $\boldsymbol{R}$ and retain its top-$r_r$ singular values: $\boldsymbol{R}_{r_r}=\boldsymbol{U}_{r_r}\boldsymbol{\Sigma}_{r_r}\boldsymbol{V}_{r_r}^T$. 
Finally, we construct the compressed weight matrix by combining the two approximations: 
\begin{align}
 \hat{\boldsymbol{W}}_r=\boldsymbol{W}_{r_i}+\boldsymbol{R}_{r_r} = \boldsymbol{U}_r\boldsymbol{\Sigma}_r\boldsymbol{V}_r^T=\hat{\boldsymbol{U}}_r\hat{\boldsymbol{V}}_r.
\end{align}
In this way, the residual-compensated compressed matrix $\hat{\boldsymbol{W}}_r$ provides a closer approximation to the original weight matrix $\boldsymbol{W}$ than the directly truncated matrix $\boldsymbol{W}_r$, as proven in the following.

\begin{lemma} \label{lemma 3.1}
    \textit{Eckart-Young-Mirsky Theorem}\textnormal{~\cite{horn2012matrix}}. \textit{Let $\boldsymbol{A}\in\mathbb{R}^{m\times n}$ be a matrix and $r\le\min(m,n)$ be a given rank. If $\boldsymbol{A}_r$ denotes the optimal rank-$r$ approximation of $\boldsymbol{A}$ by SVD, then for any matrix $\boldsymbol{B}$ of rank $r$, the following inequality holds, demonstrating that $\boldsymbol{A}_r$ is the optimal rank-$r$ approximation of $\boldsymbol{A}$:}
\begin{align}
    \big\|\boldsymbol{A}-\boldsymbol{A}_r\big\|_F^2 \le \big\|\boldsymbol{A}-\boldsymbol{B}\big\|_F^2.
\end{align}
\end{lemma}

\begin{lemma} \label{lemma 3.2}
    \textit{Let $\boldsymbol{W}'$ be the rank-$r$ approximation defined by the weighted low-rank reconstruction $\boldsymbol{W}' = SVD_r(\boldsymbol{WS})\boldsymbol{S}^{-1}$. A direct consequence of Lemma~\ref{lemma 3.1} is the following inequality:}
\begin{align}
   \big\|\boldsymbol{W}-SVD_r(\boldsymbol{W})\big\|_F^2 \le \big\|\boldsymbol{W}-\boldsymbol{W}'\big\|_F^2,
\end{align}
\end{lemma}

\textbf{Theorem 3.} \textit{If $\boldsymbol{W}_r$ is the rank-$r$ direct truncation of the original matrix $\boldsymbol{W}$, and $\hat{\boldsymbol{W}}_r$ is the residual-compensated compressed matrix, where $r=r_i + r_r$. Then, under the weight reconstruction error, ERC-SVD is superior to the direct truncation method.}
\begin{proof}
The direct truncation method extracts the first $r$ components from the $\boldsymbol{WS}$ at once. We can decompose it into two parts: the \textbf{first $r_i$ components} and the \textbf{last $r_r$ components}:
\begin{align}
   W_r = \underbrace{SVD_{r_i}(\boldsymbol{WS})\boldsymbol{S}^{-1}}_{\boldsymbol{W}_{r_i}} + \underbrace{(SVD_{r}(\boldsymbol{WS}) - SVD_{r_i}(\boldsymbol{WS}))\boldsymbol{S}^{-1}}_{\Delta}
\end{align}
For ERC-SVD, the truncation loss is given by:
\begin{gather}
   \big\|\boldsymbol{W}-\hat{\boldsymbol{W}}_r\big\|_F^2 = \big\|\boldsymbol{W}-(\boldsymbol{W}_{r_i}+\boldsymbol{R}_{r_r})\big\|_F^2 = \big\|\boldsymbol{R} - \boldsymbol{R}_{r_r}\big\|_F^2.
\end{gather}

Given that $\boldsymbol{R}_{r_r}$ is obtained by direct SVD, which is the optimal rank-$r_r$ approximation of $\boldsymbol{R}$. According to Lemma~\ref{lemma 3.1}, for any matrix $\boldsymbol{X}$ with rank $r_r$, we have:
\begin{align}
    \big\|\boldsymbol{R}-\boldsymbol{R}_{r_r}\big\|_F^2 \le \big\|\boldsymbol{R}-\boldsymbol{X}\big\|_F^2. \label{inequality11}
\end{align}

For the direct truncation method, the truncation loss is given by:
\begin{gather}
   \big\|\boldsymbol{W}-\boldsymbol{W}_r\big\|_F^2 = \big\|\boldsymbol{W}-(\boldsymbol{W}_{r_i}+\Delta)\big\|_F^2 = \big\|\boldsymbol{R} - \Delta\big\|_F^2.
\end{gather}

$\boldsymbol{R}_{r_r}$ is the optimal rank-$r_r$ approximation, whereas $\Delta$ is an approximation derived from the weighted space $\boldsymbol{WS}$. Hence, by combining Lemma~\ref{lemma 3.1} and Inequality~\ref{inequality11}, the following relationship holds:
\begin{align}
   \big\|\boldsymbol{R}-\boldsymbol{R}_{r_r}\big\|_F^2 \le \big\|\boldsymbol{R}-\Delta\big\|_F^2.
\end{align} 
By substituting these results into the original form, we obtain the final inequality:
\begin{align}
    \boxed{\big\|\boldsymbol{W}-\hat{\boldsymbol{W}}_r\big\|_F^2 \le \big\|\boldsymbol{W}-\boldsymbol{W}_r\big\|_F^2}
\end{align}
Therefore, the residual compensation strategy provides a closer approximation to the original weight matrix $\boldsymbol{W}$ than direct truncation methods.
\end{proof}

\subsection{Partial-layer Compression for SVD} \label{sec_plc}
\begin{figure}[t]
  \centering

  \begin{subfigure}[t]{0.49\linewidth}
    \centering
    \includegraphics[width=\linewidth]{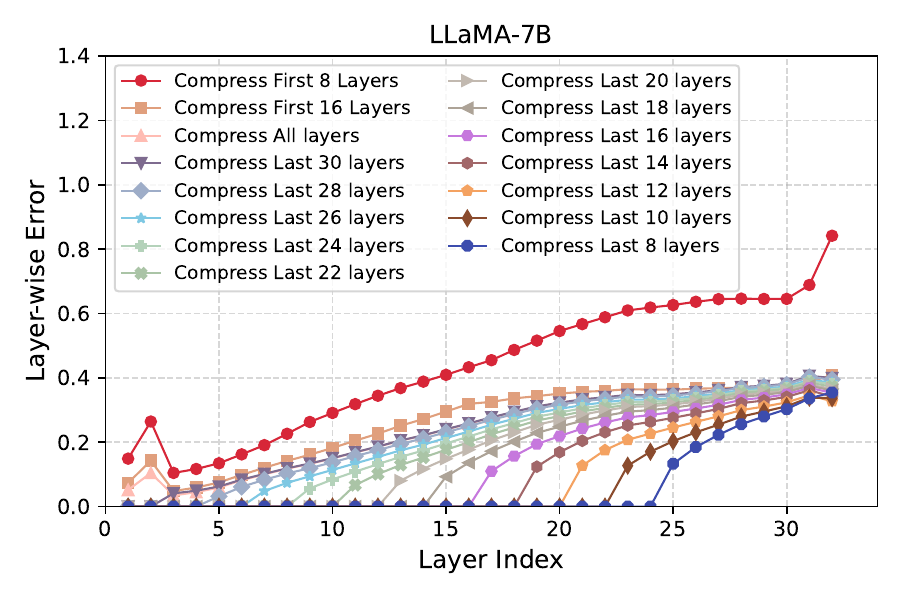}
  \end{subfigure}
  % \hfill
  \begin{subfigure}[t]{0.49\linewidth}
    \centering
    \includegraphics[width=\linewidth]{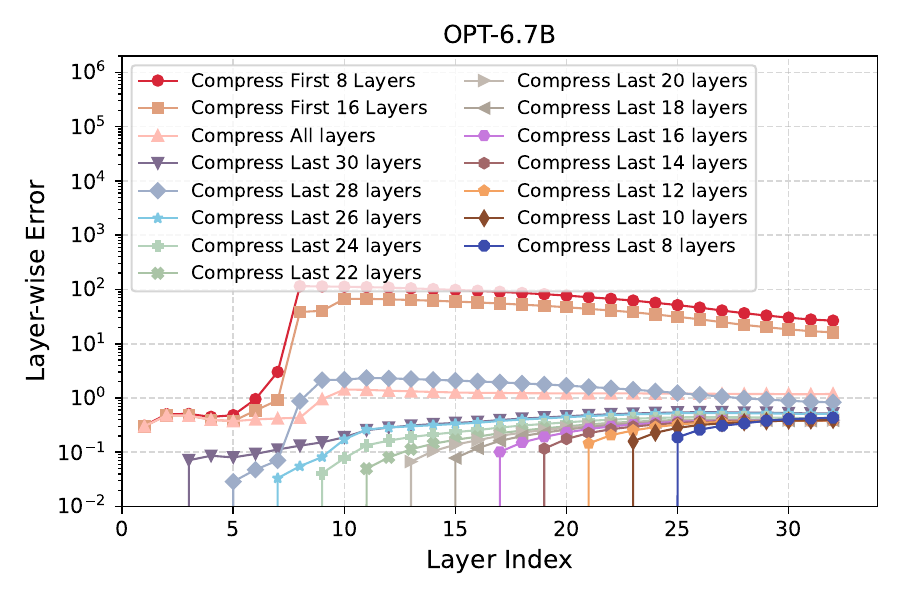}
  \end{subfigure}

  \caption{
    Layer-wise error comparison between the original model, LLaMA-7B, and OPT-6.7B compressed by ERC-SVD with different layer selection strategies on WikiText-2. The overall compression ratio is 20\%, and all layer selection strategies strictly adhere to the compression constraint.
  }
  \label{layer-wise_error_comparison_llm}
\end{figure}

Previous works~\cite{yuan2023asvd,wang2024svd,wang2024basis,li2025adasvd} compress all model layers even if they assign layer-specific ratios based on their relative importance, which often leads to a high layer-wise error, resulting in noticeable performance degradation. We compare the layer-wise error of LLMs across four families with different layer selection strategies; the results of LLaMA-7B and OPT-6.7B are shown in Figure~\ref{layer-wise_error_comparison_llm}, the results of other models are presented in Figure~\ref{layer-wise_error_comparison_llm-1}. There is a significant error propagation across multiple LLM families, where the error progressively accumulates layer by layer during the forward pass. 
This phenomenon is particularly pronounced when compressing only the first few layers. 
A higher layer compression ratio must be applied to these layers, with the constraint of the overall compression ratio.
Compressing the model at such an early stage introduces substantial approximation errors, which are then propagated through the remaining layers.
Although the later layers are left uncompressed, the forward pass carries the accumulated error through the network. Consequently, in all evaluated LLM families, these early compressed layers exhibit the highest layer-wise error across model layers. 
In contrast, when only compressing the last few layers, the earlier layers introduce zero error. Although the relatively high compression ratio leads to slightly faster error accumulation, the overall error remains significantly lower than that of compressing all layers.

Interestingly, although the final-layer error converges to a narrow range regardless of how many the last layers are compressed, differences within this range still affect model performance. As illustrated in the left figure of Figure~\ref{fig:teaser}, there is a strong correlation between the final-layer error and the average zero-shot accuracy. Motivated by this observation, we select the number of last layers to compress by minimizing the final-layer error, and the layer compression ratio $R_l$ must satisfy the following mathematical relation to ensure a fair comparison:
\begin{align}
   R_l = (N \cdot R_o)/k, k \in \left\{ 1, 2, \dots, N-1 \;\middle|\; R_l< 1\right\},
\end{align} 
where $N$ is the number of model layers, $R_o$ represents the overall compression ratio.
As shown in the middle figure of Figure~\ref{fig:teaser}, compared to SVD-LLM~\cite{wang2024svd}, our strategy significantly reduces layer-wise error across all model layers and mitigates error propagation.

\section{Experiments} \label{sec:experiments}
\subsection{Setups}

\paragraph{Evaluation Details.} We compare our method against five baselines without re-training: conventional SVD, ASVD~\cite{yuan2023asvd}, SVD-LLM~\cite{wang2024svd}, Basis Sharing~\cite{wang2024basis}, and AdaSVD~\cite{li2025adasvd}. We evaluate ERC-SVD on ten models spanning five LLM families: LLaMA-2-7B~\cite{touvron2023llama2}, LLaMA-13B~\cite{touvron2023llama}, LLaMA-2-13B~\cite{touvron2023llama2}, LLaMA-30B~\cite{touvron2023llama}, OPT-6.7B~\cite{zhang2022opt}, OPT-13B~\cite{zhang2022opt}, OPT-30B~\cite{zhang2022opt}, Mistral-7B~\cite{jiang2023mistral}, Vicuna-7B~\cite{chiang2023vicuna}, Qwen-3-8B~\cite{yang2025qwen3}. 
For language modeling, we use three benchmark datasets: WikiText-2~\cite{merity2016pointer}, PTB~\cite{marcus1993building}, and C4~\cite{raffel2020exploring}.
For zero-shot reasoning and understanding, we evaluate on seven tasks within the \textit{LM-Evaluation-Harness} framework: OpenbookQA~\cite{mihaylov2018can}, WinoGrande~\cite{sakaguchi2021winogrande}, HellaSwag~\cite{zellers2019hellaswag}, PIQA~\cite{bisk2020piqa}, MathQA~\cite{amini2019mathqa}, ARC\_e and ARC\_c~\cite{clark2018think}. Moreover, to demonstrate the generality and effectiveness of ERC-SVD, we further extend our evaluation to vision-language models (VLMs), LLaVA~\cite{liu2024improved}.

\paragraph{Implementation Details.} \label{implementation_details}
We follow the protocols of ASVD, SVD-LLM, and Basis Sharing, randomly selecting 256 calibration samples with a sequence length of 2048 from WikiText-2. We then apply matrix scaling prior to performing SVD truncation. All methods are implemented with PyTorch and Transformers on NVIDIA A100 GPUs. 

\subsection{Results}
We evaluate the overall performance of ERC-SVD across four dimensions: \ding{172} \textbf{Effectiveness under different compression ratios} (ranging from 20\% to 60\%), \ding{173} \textbf{Scalability to larger-scale models}, \ding{174} \textbf{Generalizability across diverse LLM families}, and \ding{175} \textbf{Performance on VLMs}, including quantitative and qualitative results, are provided in Table~\ref{some_benchmark} and Appendix~\ref{ressvd:vlm}, respectively.
% \FloatBarrier
\begin{table}[t]
\centering
\caption{Overall performance of LLaMA-2-7B~\cite{touvron2023llama2} compressed by ERC-SVD and baselines under 20\% to 60\% compression ratios (``\textsc{Ratio}''), including performance on three language modeling datasets (measured by perplexity (\textcolor{blue}{$\downarrow$})) and zero-shot performance on seven common sense reasoning datasets (measured by individual and average accuracy (\textcolor{green!60!black}{$\uparrow$})). The best results are marked in \textbf{bold}. NaN denotes evaluation failure due to numerical instability. $*$ refers to results derived from the original paper. - means that results are not available.}
% \vspace{1mm}
% \vspace{-5pt}
\label{llama2_result_table}
\small
\resizebox{\textwidth}{!}{
\begin{tabular}{c|c|ccc|cccccccc}
    \toprule
     \textsc{Ratio} & \cellcolor{white} \textsc{Method} & \cellcolor{white} WikiText-2\textcolor{blue}{$\downarrow$} & \cellcolor{white} PTB\textcolor{blue}{$\downarrow$} & \cellcolor{white} C4\textcolor{blue}{$\downarrow$} & \cellcolor{white} Openb.$\uparrow$  & \cellcolor{white} ARC\_e$\uparrow$ & \cellcolor{white} WinoG.$\uparrow$ & \cellcolor{white} HellaS.$\uparrow$ & \cellcolor{white} ARC\_c$\uparrow$ & \cellcolor{white} PIQA$\uparrow$ & \cellcolor{white} MathQA$\uparrow$ & \cellcolor{white} Average\textcolor{green!60!black}{$\uparrow$}\\
    \midrule
    \textcolor{gray}{-} & \cellcolor{white} \textcolor{gray}{Original} & \cellcolor{white} \textcolor{gray}{5.47} & \cellcolor{white} \textcolor{gray}{26.84} & \cellcolor{white} \textcolor{gray}{7.28} & \cellcolor{white} \textcolor{gray}{0.31} & \cellcolor{white} \textcolor{gray}{0.69} & \cellcolor{white} \textcolor{gray}{0.67} & \cellcolor{white} \textcolor{gray}{0.56} & \cellcolor{white} \textcolor{gray}{0.40} & \cellcolor{white} \textcolor{gray}{0.78} & \cellcolor{white} \textcolor{gray}{0.27} & \cellcolor{white} \textcolor{gray}{0.53}\\
    \midrule
        \multirow{5}{*}{20\%}  & \cellcolor{white} SVD & \cellcolor{white} 18208.79 & \cellcolor{white} 59320.78 & \cellcolor{white} 27131.56 & \cellcolor{white} 0.14 & \cellcolor{white} 0.26 & \cellcolor{white} 0.49 & \cellcolor{white} 0.25 & \cellcolor{white} 0.22 & \cellcolor{white} 0.52 & \cellcolor{white} 0.20 & \cellcolor{white} 0.30\\

                               & \cellcolor{white} ASVD & \cellcolor{white} 9.56 & \cellcolor{white} 120.74 & \cellcolor{white} \textbf{12.85} & \cellcolor{white} \underline{0.25} & \cellcolor{white} 0.59 & \cellcolor{white} \underline{0.61} & \cellcolor{white} \underline{0.46} & \cellcolor{white} \underline{0.32} & \cellcolor{white} \underline{0.72} & \cellcolor{white} \underline{0.24} & \cellcolor{white} \underline{0.45}\\
              
                               & \cellcolor{white} SVD-LLM & \cellcolor{white} 8.37 & \cellcolor{white} 139.68 & \cellcolor{white} 20.18 & \cellcolor{white} 0.23 & \cellcolor{white} 0.50 & \cellcolor{white} 0.59 & \cellcolor{white} 0.39 & \cellcolor{white} 0.26 & \cellcolor{white} 0.65 & \cellcolor{white} 0.23 & \cellcolor{white} 0.41\\

                               & \cellcolor{white} Basis Sharing & \cellcolor{white} \underline{7.77} & \cellcolor{white} \underline{60.00} & \cellcolor{white} 16.22 & \cellcolor{white} 0.24 & \cellcolor{white} \underline{0.60} & \cellcolor{white} \underline{0.61} & \cellcolor{white} 0.43 & \cellcolor{white} 0.31 & \cellcolor{white} 0.69 & \cellcolor{white} \underline{0.24} &  \cellcolor{white} \underline{0.45} \\
        \cmidrule{2-13}      
                               & \cellcolor{cyan!10} \textbf{ERC-SVD} & \cellcolor{cyan!10} \textbf{7.63}  & \cellcolor{cyan!10} \textbf{45.37}  & \cellcolor{cyan!10} \underline{14.73} & \cellcolor{cyan!10} \textbf{0.28} & \cellcolor{cyan!10} \textbf{0.61} & \cellcolor{cyan!10} \textbf{0.65} & \cellcolor{cyan!10} \textbf{0.50} & \cellcolor{cyan!10} \textbf{0.35} & \cellcolor{cyan!10} \textbf{0.73} & \cellcolor{cyan!10} \textbf{0.26} & \cellcolor{cyan!10} \textbf{0.48} \\

    \midrule
        \multirow{5}{*}{30\%}  & \cellcolor{white} SVD & \cellcolor{white} 30373.39 & \cellcolor{white} 48930.94 & \cellcolor{white} 36905.54 & \cellcolor{white} 0.12 & \cellcolor{white} 0.25 & \cellcolor{white} 0.49 & \cellcolor{white} 0.25 & \cellcolor{white} 0.22 & \cellcolor{white} 0.51 & \cellcolor{white} 0.21 & \cellcolor{white} 0.29\\

                               & \cellcolor{white} ASVD & \cellcolor{white} 984.03 & \cellcolor{white} NaN & \cellcolor{white} NaN & \cellcolor{white} 0.15 & \cellcolor{white} 0.27 & \cellcolor{white} 0.51 & \cellcolor{white} 0.26 & \cellcolor{white} 0.22 & \cellcolor{white} 0.53 & \cellcolor{white} 0.20 & \cellcolor{white} 0.31\\
        
                               & \cellcolor{white} SVD-LLM & \cellcolor{white} 10.66 & \cellcolor{white} 292.90 & \cellcolor{white} 34.96 & \cellcolor{white} 0.21 & \cellcolor{white} 0.42 & \cellcolor{white} 0.55 & \cellcolor{white} 0.34 & \cellcolor{white} 0.22 & \cellcolor{white} 0.60 & \cellcolor{white} \underline{0.23} & \cellcolor{white} 0.37\\

                               & \cellcolor{white} Basis Sharing & \cellcolor{white} \textbf{9.69} & \cellcolor{white} \underline{101.22} & \cellcolor{white} \underline{25.45} & \cellcolor{white} \underline{0.23} & \cellcolor{white} \textbf{0.54} & \cellcolor{white} \underline{0.59} & \cellcolor{white} \underline{0.38} & \cellcolor{white} \underline{0.27} & \cellcolor{white} \underline{0.64} & \cellcolor{white} \underline{0.23} & \cellcolor{white} \underline{0.41} \\
        \cmidrule{2-13}      
                               & \cellcolor{cyan!10} \textbf{ERC-SVD} & \cellcolor{cyan!10} \underline{10.32} & \cellcolor{cyan!10} \textbf{73.04} & \cellcolor{cyan!10} \textbf{23.28} & \cellcolor{cyan!10} \textbf{0.23} & \cellcolor{cyan!10} \underline{0.51} & \cellcolor{cyan!10} \textbf{0.62} & \cellcolor{cyan!10} \textbf{0.42} & \cellcolor{cyan!10} \textbf{0.29} & \cellcolor{cyan!10} \textbf{0.68} & \cellcolor{cyan!10} \textbf{0.24} & \cellcolor{cyan!10} \textbf{0.43} \\
    \midrule
        \multirow{6}{*}{40\%}  & \cellcolor{white} SVD & \cellcolor{white} 39524.00 & \cellcolor{white} 68829.98 & \cellcolor{white} 56518.95 & \cellcolor{white} 0.13 & \cellcolor{white} 0.26 & \cellcolor{white} 0.50 & \cellcolor{white} 0.25 & \cellcolor{white} 0.21 & \cellcolor{white} 0.52 & \cellcolor{white} 0.18 & \cellcolor{white} 0.29\\

                               & \cellcolor{white} ASVD & \cellcolor{white} NaN & \cellcolor{white} NaN & \cellcolor{white} NaN & \cellcolor{white} 0.15 & \cellcolor{white} 0.25 & \cellcolor{white} 0.50 & \cellcolor{white} 0.26 & \cellcolor{white} 0.22 & \cellcolor{white} 0.52 & \cellcolor{white} 0.18 & \cellcolor{white} 0.30\\
        
                               & \cellcolor{white} SVD-LLM & \cellcolor{white} 16.11 & \cellcolor{white} 717.34 & \cellcolor{white} 61.96 & \cellcolor{white} 0.16 & \cellcolor{white} 0.35 & \cellcolor{white} 0.55 & \cellcolor{white} 0.30 & \cellcolor{white} 0.20 & \cellcolor{white} 0.57 & \cellcolor{white} \underline{0.23} & \cellcolor{white} 0.34\\

                               & \cellcolor{white} Basis Sharing & \cellcolor{white} \underline{14.66} & \cellcolor{white} \underline{207.49} & \cellcolor{white} \underline{47.41} & \cellcolor{white} 0.18 & \cellcolor{white} \underline{0.43} & \cellcolor{white} \underline{0.57} & \cellcolor{white} \underline{0.33} & \cellcolor{white} 0.22 & \cellcolor{white} \underline{0.61} & \cellcolor{white} \underline{0.23} & \cellcolor{white} \underline{0.37} \\

                               & \cellcolor{white} AdaSVD$^{*}$ & \cellcolor{white} 14.76 & \cellcolor{white} 304.62 & \cellcolor{white} 56.98 & \cellcolor{white} \underline{0.19} & \cellcolor{white} 0.41 & \cellcolor{white} \textbf{0.58} & \cellcolor{white} 0.32 & \cellcolor{white} \underline{0.23} & \cellcolor{white} 0.58 & \cellcolor{white} - & \cellcolor{white} -\\
        \cmidrule{2-13}    
                               & \cellcolor{cyan!10} \textbf{ERC-SVD} & \cellcolor{cyan!10} \textbf{14.17} & \cellcolor{cyan!10} \textbf{136.32} & \cellcolor{cyan!10} \textbf{43.19} & \cellcolor{cyan!10} \textbf{0.20} & \cellcolor{cyan!10} \textbf{0.43} & \cellcolor{cyan!10} \underline{0.57} & \cellcolor{cyan!10} \textbf{0.35} & \cellcolor{cyan!10} \textbf{0.24} & \cellcolor{cyan!10} \textbf{0.63} & \cellcolor{cyan!10} \textbf{0.23} & \cellcolor{cyan!10} \textbf{0.38} \\
    \midrule
        \multirow{6}{*}{50\%}  & \cellcolor{white} SVD & \cellcolor{white} 53405.48 & \cellcolor{white} 39023.05 & \cellcolor{white} 58547.82 & \cellcolor{white} 0.15 & \cellcolor{white} 0.25 & \cellcolor{white} 0.48 & \cellcolor{white} 0.25 & \cellcolor{white} \underline{0.22} & \cellcolor{white} 0.53 & \cellcolor{white} 0.18 & \cellcolor{white} 0.29\\

                               & \cellcolor{white} ASVD & \cellcolor{white} NaN & \cellcolor{white} NaN & \cellcolor{white} NaN & \cellcolor{white} 0.13 & \cellcolor{white} 0.26 & \cellcolor{white} 0.50 & \cellcolor{white} 0.25 & \cellcolor{white} \textbf{0.23} & \cellcolor{white} 0.50 & \cellcolor{white} 0.20 & \cellcolor{white} 0.30\\
              
                               & \cellcolor{white} SVD-LLM & \cellcolor{white} 27.19 & \cellcolor{white} 1775.52 & \cellcolor{white} 129.71 & \cellcolor{white} \underline{0.14} & \cellcolor{white} 0.30 & \cellcolor{white} 0.50 & \cellcolor{white} 0.28 & \cellcolor{white} 0.20 & \cellcolor{white} 0.54 & \cellcolor{white} 0.23 & \cellcolor{white} 0.31\\

                               & \cellcolor{white} Basis Sharing & \cellcolor{white} \textbf{23.78} & \cellcolor{white} 689.03 & \cellcolor{white} 115.84 & \cellcolor{white} 0.15 & \cellcolor{white} \underline{0.34} & \cellcolor{white} \underline{0.54} & \cellcolor{white} \underline{0.29} & \cellcolor{white} 0.20 & \cellcolor{white} \underline{0.56} & \cellcolor{white} \textbf{0.23} & \cellcolor{white} \underline{0.33} \\

                               & \cellcolor{white} AdaSVD$^{*}$ & \cellcolor{white} 25.58 & \cellcolor{white} \underline{593.14} & \cellcolor{white} \underline{113.84}  & \cellcolor{white} \textbf{0.15} & \cellcolor{white} \underline{0.34} & \cellcolor{white} \underline{0.54} & \cellcolor{white} \underline{0.29} & \cellcolor{white} 0.20 & \cellcolor{white} \underline{0.56} & \cellcolor{white} - & \cellcolor{white} -\\
        \cmidrule{2-13}      
                               & \cellcolor{cyan!10} \textbf{ERC-SVD} & \cellcolor{cyan!10} \underline{24.26} & \cellcolor{cyan!10} \textbf{286.24} & \cellcolor{cyan!10} \textbf{100.34} & \cellcolor{cyan!10} \underline{0.14} & \cellcolor{cyan!10} \textbf{0.35} & \cellcolor{cyan!10} \textbf{0.55} & \cellcolor{cyan!10} \textbf{0.31} & \cellcolor{cyan!10} \underline{0.22} & \cellcolor{cyan!10} \textbf{0.59} & \cellcolor{cyan!10} \underline{0.22} & \cellcolor{cyan!10} \textbf{0.34} \\
    \midrule
        \multirow{6}{*}{60\%}  & \cellcolor{white} SVD & \cellcolor{white} 65240.23 & \cellcolor{white} 79002.21 & \cellcolor{white} 70659.74 & \cellcolor{white} \underline{0.14} & \cellcolor{white} 0.25 & \cellcolor{white} 0.50 & \cellcolor{white} 0.25 & \cellcolor{white} 0.23 & \cellcolor{white} 0.52 & \cellcolor{white} 0.19 & \cellcolor{white} \underline{0.30}\\

                               & \cellcolor{white} ASVD & \cellcolor{white} NaN & \cellcolor{white} 19581.17 & \cellcolor{white} NaN & \cellcolor{white} \textbf{0.15} & \cellcolor{white} 0.25 & \cellcolor{white} 0.50 & \cellcolor{white} 0.25 & \cellcolor{white} \textbf{0.23} & \cellcolor{white} 0.52 & \cellcolor{white} 0.12 & \cellcolor{white} 0.29\\
        
                               & \cellcolor{white} SVD-LLM & \cellcolor{white} \underline{54.19} & \cellcolor{white} 3442.74 & \cellcolor{white} \underline{263.02} & \cellcolor{white} \underline{0.14} & \cellcolor{white} 0.26 & \cellcolor{white} 0.50 & \cellcolor{white} \underline{0.27} & \cellcolor{white} 0.20 & \cellcolor{white} \underline{0.53} & \cellcolor{white} \underline{0.21} & \cellcolor{white} \underline{0.30}\\

                               & \cellcolor{white} Basis Sharing & \cellcolor{white} \textbf{50.59} & \cellcolor{white} \underline{2055.63} & \cellcolor{white} 299.36 & \cellcolor{white} \underline{0.14} & \cellcolor{white} \underline{0.29} & \cellcolor{white} \underline{0.51} & \cellcolor{white} \underline{0.27} & \cellcolor{white} 0.20 & \cellcolor{white} 0.52 & \cellcolor{white} \underline{0.21} & \cellcolor{white} \underline{0.30} \\

                               & \cellcolor{white} AdaSVD$^{*}$ & \cellcolor{white} 60.08 & \cellcolor{white} 2137.28 & \cellcolor{white} 294.26 & \cellcolor{white} 0.12 & \cellcolor{white} \underline{0.27} & \cellcolor{white} 0.50 & \cellcolor{white} 0.27 & \cellcolor{white} 0.20 & \cellcolor{white} \underline{0.53} & \cellcolor{white} - & \cellcolor{white} -\\
        \cmidrule{2-13}    
                               & \cellcolor{cyan!10} \textbf{ERC-SVD} & \cellcolor{cyan!10} 68.59 & \cellcolor{cyan!10} \textbf{991.48} & \cellcolor{cyan!10} \textbf{255.70} & \cellcolor{cyan!10} 0.13 & \cellcolor{cyan!10} \textbf{0.29} & \cellcolor{cyan!10} \textbf{0.52} & \cellcolor{cyan!10} \textbf{0.28} & \cellcolor{cyan!10} \underline{0.21} & \cellcolor{cyan!10} \textbf{0.55} & \cellcolor{cyan!10} \textbf{0.22} & \cellcolor{cyan!10} \textbf{0.31} \\
    \bottomrule
\end{tabular}
}
\end{table}

\paragraph{Performance under different compression ratios.} 
We evaluate the performance of LLaMA-2-7B~\cite{touvron2023llama2} compressed by ERC-SVD, conventional SVD, and existing post-training baselines under compression ratios ranging from 20\% to 60\% across ten benchmark datasets. 
The results for LLaMA-2-7B are presented in Table~\ref{llama2_result_table}. 
Across three language modeling datasets, WikiText-2~\cite{merity2016pointer}, PTB~\cite{marcus1993building}, and C4~\cite{raffel2020exploring}, ERC-SVD consistently outperforms all baselines across most evaluated compression ratios, with only slight suboptimalities observed in specific cases. 
In particular, on PTB and C4, the improvements are more pronounced, suggesting that ERC-SVD exhibits stronger generalization capability. In addition, across seven common sense reasoning datasets, ERC-SVD outperforms the strongest existing baseline on most datasets, achieving the highest average accuracy across all evaluated compression ratios and further demonstrating its robustness and overall effectiveness.

\begin{figure}[h]
    \centering
    \begin{minipage}[t]{0.49\textwidth}
        \vspace{0pt}
        \normalsize
        \paragraph{Performance on Multiple LLM families.}
        To evaluate the generalization ability of ERC-SVD, we apply it to three LLMs from different families, including OPT-6.7B~\cite{zhang2022opt}, Mistral-7B~\cite{jiang2023mistral}, and Vicuna-7B~\cite{chiang2023vicuna}. 
        As shown in Table~\ref{different_llm_families}, under 30\% compression ratio, ERC-SVD consistently outperforms all baselines on three language modeling benchmarks across these architectures. 
        Notably, the largest overall improvements are observed on Mistral-7B, with perplexity reductions of 71\% on WikiText-2, 45\% on PTB, and 46\% on C4. 
        We reproduce ASVD and SVD-LLM using their public codebases. While ASVD fails in certain cases due to numerical instability (denoted as NaN in the table), ERC-SVD consistently maintains stable and reliable performance. Additional zero-shot accuracy results comparison about these LLMs are provided in Table~\ref{multiple-llm-acc}.
    \end{minipage}%
    \hfill
    \begin{minipage}[t]{0.455\textwidth}
        \vspace{0pt}
        \centering
        \captionof{table}{Perplexity (\textcolor{blue}{$\downarrow$}) of different LLM structures under 30\% compression ratio.}
        % NaN denotes evaluation failure as numerical instability.
        % \vspace{-6pt}
        \vspace{-8pt}
        \label{different_llm_families}
        \small
        \resizebox{\linewidth}{!}{%
        \begin{tabular}{c|c|ccc}
        \toprule
        \textsc{Model} & \textsc{Method} & WikiText-2\textcolor{blue}{$\downarrow$} & PTB\textcolor{blue}{$\downarrow$} & C4\textcolor{blue}{$\downarrow$} \\
        \midrule
            \multirow{4}{*}{OPT-6.7B} & \cellcolor{white} SVD & \cellcolor{white} 116067.28 & \cellcolor{white} 86760.50 & \cellcolor{white} 168165.89\\
    
                                              & \cellcolor{white} ASVD & \cellcolor{white} \underline{26.67} & \cellcolor{white} 71.36 & \cellcolor{white} 44.51\\
                  
                                              & \cellcolor{white} SVD-LLM & \cellcolor{white} 28.03 & \cellcolor{white} \underline{37.46} & \cellcolor{white} \underline{40.35} \\
        \cmidrule{2-5} 
                                              & \cellcolor{cyan!10} \textbf{ERC-SVD} & \cellcolor{cyan!10} \textbf{17.10} (\textcolor{blue}{$\downarrow$36\%}) & \cellcolor{cyan!10} \textbf{27.24} (\textcolor{blue}{$\downarrow$27\%}) & \cellcolor{cyan!10} \textbf{38.40} (\textcolor{blue}{$\downarrow$5\%}) \\
        \midrule
            \multirow{4}{*}{Mistral-7B} & \cellcolor{white} SVD & \cellcolor{white} 59569.54 & \cellcolor{white} 57830.63 & \cellcolor{white} 78168.24\\
    
                                                & \cellcolor{white} ASVD & \cellcolor{white} 221.66 & \cellcolor{white} 927.15 & \cellcolor{white} 266.04\\
                  
                                                & \cellcolor{white} SVD-LLM & \cellcolor{white} \underline{48.94} & \cellcolor{white} \underline{193.22} & \cellcolor{white} \underline{56.55}\\
        \cmidrule{2-5}  
                                                & \cellcolor{cyan!10} \textbf{ERC-SVD} & \cellcolor{cyan!10} \textbf{14.09} (\textcolor{blue}{$\downarrow$71\%}) & \cellcolor{cyan!10} \textbf{105.37} (\textcolor{blue}{$\downarrow$45\%}) & \cellcolor{cyan!10} \textbf{30.72} (\textcolor{blue}{$\downarrow$46\%})\\
        \midrule
            \multirow{4}{*}{Vicuna-7B} & \cellcolor{white} SVD & \cellcolor{white} 24835.33 & \cellcolor{white} 24510.90 & \cellcolor{white} 29368.55\\
    
                                               & \cellcolor{white} ASVD & \cellcolor{white} 106.32 & \cellcolor{white} NaN & \cellcolor{white} NaN\\
                  
                                               & \cellcolor{white} SVD-LLM & \cellcolor{white} \underline{12.42} & \cellcolor{white} \underline{104.27} & \cellcolor{white} \underline{39.55}\\
        \cmidrule{2-5}
                                               & \cellcolor{cyan!10} \textbf{ERC-SVD} & \cellcolor{cyan!10} \textbf{11.57} (\textcolor{blue}{$\downarrow$7\%}) & \cellcolor{cyan!10} \textbf{69.28} (\textcolor{blue}{$\downarrow$34\%}) & \cellcolor{cyan!10} \textbf{27.24} (\textcolor{blue}{$\downarrow$31\%})\\
        \bottomrule
    \end{tabular}
    }
    \vspace{-2pt}
    \captionof{table}{Perplexity (\textcolor{blue}{$\downarrow$}) of larger-scale LLMs under 20\% compression ratio on PTB~\cite{marcus1993building}.}
    \vspace{1pt}
    \label{larger_scale_llms}
    \small
    \resizebox{\linewidth}{!}{
    \begin{tabular}{c|cc|c|cc}
        \toprule
        \textsc{Method} & LLaMA-13B & LLaMA-30B & LLaMA-2-13B & OPT-13B & OPT-30B \\
        \midrule
        SVD       & 1878.04 & 555.55 & 5464.57 & 1552.55 & 250.49 \\
        ASVD      & 12.42   & 31.36  & \underline{81.00}   & 36.47   & 30.95  \\
        SVD-LLM   & \underline{12.17}   & \underline{9.10}   & 88.13   & \underline{14.86}   & \underline{12.94}  \\
        \cmidrule{1-6}
        \rowcolor{cyan!10} \textbf{ERC-SVD} 
                   & \textbf{9.70} & \textbf{8.41} & \textbf{66.47} & \textbf{13.22} & \textbf{12.89} \\
        \bottomrule
    \end{tabular}
        }
    \end{minipage}
\end{figure}

\paragraph{Performance on larger-scale LLMs.} 
To examine the scalability of ERC-SVD, we evaluate its performance on larger-scale LLMs from two representative families: LLaMA and OPT (13B and 30B).
% (including LLaMA-13B, LLaMA-30B, and LLaMA-2-13B) and OPT (OPT-13B and OPT-30B). 
As presented in Table~\ref{larger_scale_llms}, ERC-SVD consistently achieves superior performance over existing baselines under 20\% compression ratio, demonstrating robust effectiveness across varying model scales. Additional zero-shot accuracy results about these larger-scale LLMs are provided in Table~\ref{larger-llm-acc}.

\begin{table}[ht]
\centering
\caption{Ablation results of REC and PLC on LLaMA-2-7B under 30\% and 40\% compression ratios.}
% \vspace{1mm}
% \vspace{-5pt}
\label{ablation_study}
\small
\resizebox{\textwidth}{!}{
\begin{tabular}{c|c|cc|c|cccccccc}
    \toprule
     \textsc{Ratio} & \textsc{Method} & REC & PLC & C4\textcolor{blue}{$\downarrow$} & Openb.$\uparrow$  & ARC\_e$\uparrow$ & WinoG.$\uparrow$ & HellaS.$\uparrow$ & ARC\_c$\uparrow$ & PIQA$\uparrow$ & MathQA$\uparrow$ & Average\textcolor{green!60!black}{$\uparrow$}\\
    \midrule
        \multirow{5}{*}{30\%} 
                               & ASVD & \cellcolor{white} - & \cellcolor{white} - & \cellcolor{white} NaN & \cellcolor{white} 0.15 & \cellcolor{white} 0.27 & \cellcolor{white} 0.51 & \cellcolor{white} 0.26 & \cellcolor{white} 0.22 & \cellcolor{white} 0.53 & \cellcolor{white} 0.20 & \cellcolor{white} 0.31\\
        
                               & SVD-LLM & \cellcolor{white} - & \cellcolor{white} - & \cellcolor{white} 34.96 & \cellcolor{white} 0.21 & \cellcolor{white} 0.42 & \cellcolor{white} 0.55 & \cellcolor{white} 0.34 & \cellcolor{white} 0.22 & \cellcolor{white} 0.60 & \cellcolor{white} 0.23 & \cellcolor{white} 0.37\\

        \cmidrule{2-13}
        & \multirow{3}{*}{\textbf{ERC-SVD}}
                               & \cellcolor{white} \ding{51} & \cellcolor{white} \ding{55} & \cellcolor{white} 30.68 & \cellcolor{white} 0.22 & \cellcolor{white} 0.42 & \cellcolor{white} 0.58 & \cellcolor{white} 0.35 & \cellcolor{white} 0.23 & \cellcolor{white} 0.61 & \cellcolor{white} 0.24 & \cellcolor{white} 0.38\\

                               & & \cellcolor{white} \ding{55} & \cellcolor{white} \ding{51} & \cellcolor{white} 24.77 & \cellcolor{white} 0.22 & \cellcolor{white} 0.48 & \cellcolor{white} 0.60 & \cellcolor{white} 0.38 & \cellcolor{white} 0.25 & \cellcolor{white} 0.65 & \cellcolor{white} 0.22 & \cellcolor{white} 0.40\\
                               
                               & & \cellcolor{cyan!10} \ding{51} & \cellcolor{cyan!10} \ding{51} & \cellcolor{cyan!10} \textbf{23.28}  & \cellcolor{cyan!10} 0.23 & \cellcolor{cyan!10} 0.51 & \cellcolor{cyan!10} 0.62 & \cellcolor{cyan!10} 0.42 & \cellcolor{cyan!10} 0.29 & \cellcolor{cyan!10} 0.68 & \cellcolor{cyan!10} 0.24 & \cellcolor{cyan!10} \textbf{0.43} \\
    \bottomrule
\end{tabular}
}
\end{table}

\begin{table}[ht]
\centering
\caption{Perplexity (\textcolor{blue}{$\downarrow$}) and zero-shot individual and average accuracy (\textcolor{green!60!black}{$\uparrow$}) of LLaMA-2-7B 30\% compressed by ERC-SVD with different residual compensation factors ($\beta$).}
\label{tab:residual_compensation_factor}
% \vspace{1mm}
% \vspace{-5pt}
\small
\resizebox{\textwidth}{!}{
\begin{tabular}{c|c|cccccccc}
    \toprule
     $\beta$ & WikiText-2\textcolor{blue}{$\downarrow$} & Openb.$\uparrow$  & ARC\_e$\uparrow$ & WinoG.$\uparrow$ & HellaS.$\uparrow$ & ARC\_c$\uparrow$ & PIQA$\uparrow$ & MathQA$\uparrow$ & Average\textcolor{green!60!black}{$\uparrow$}\\
    \midrule
        \cellcolor{white} 0.025 & \cellcolor{white} 10.38 & \cellcolor{white} 0.23 & \cellcolor{white} 0.51 & \cellcolor{white} 0.62 & \cellcolor{white} 0.41 & \cellcolor{white} 0.28 & \cellcolor{white} 0.67 & \cellcolor{white} 0.23 & \cellcolor{white} 0.42 \\
              
        \cellcolor{cyan!10} 0.050 & \cellcolor{cyan!10} 10.32 & \cellcolor{cyan!10} 0.23 & \cellcolor{cyan!10} 0.51 & \cellcolor{cyan!10} 0.62 & \cellcolor{cyan!10} 0.42 & \cellcolor{cyan!10} 0.29 & \cellcolor{cyan!10} 0.68 & \cellcolor{cyan!10} 0.24 & \cellcolor{cyan!10} \textbf{0.43} \\

        \cellcolor{white} 0.075 & \cellcolor{white} \textbf{10.11} & \cellcolor{white} 0.22 & \cellcolor{white} 0.49 & \cellcolor{white} 0.61 & \cellcolor{white} 0.39 & \cellcolor{white} 0.28 & \cellcolor{white} 0.66 & \cellcolor{white} 0.23 & \cellcolor{white} 0.41 \\

        \cellcolor{white} 0.010 & \cellcolor{white} 10.46 & \cellcolor{white} 0.23 & \cellcolor{white} 0.47 & \cellcolor{white} 0.60 & \cellcolor{white} 0.38 & \cellcolor{white} 0.26 & \cellcolor{white} 0.66 & \cellcolor{white} 0.23 & \cellcolor{white} 0.40 \\
    \bottomrule
\end{tabular}
}
\end{table}

\subsection{Ablation Study}
We present several ablation studies to assess the robustness of ERC-SVD. 
\ding{172} \textbf{Effectiveness of residual compensation (REC) and partial-layer compression (PLC):} We assess the individual contributions of REC and PLC.
\ding{173} \textbf{Impact of residual compensation factor:} We conduct experiments to examine how the choice of $\beta$ influences the performance.
\ding{174} \textbf{Impact of calibration data:} We analyze the effects of calibration dataset selection on compressed model performance in Appendix~\ref{calibration-data}.

\paragraph{Effectiveness of residual compensation and partial-layer compression.} 
We present results on C4~\cite{raffel2020exploring} and seven zero-shot reasoning and understanding tasks, as shown in Table~\ref{ablation_study}. It can be observed that either applying REC or PLC alone has already improved performance. Combining them leads to further improvements, making the performance gap even more significant. Results of other compression ratios are presented in Table~\ref{more_ablation_study}.

\paragraph{Impact of residual compensation factor.} 
\label{residual_compensation_factor}
The residual compensation factor $\beta$ is a hyperparameter in this work, we set $\beta = 0.05$ across all experiments. To examine its impact on model performance, we compress LLaMA-2-7B under 30\% compression ratio while varying the value of $\beta$. The results in Table~\ref{tab:residual_compensation_factor} show that the compressed models achieve comparable performance across different $\beta$ values, demonstrating the robustness of our method with respect to this hyperparameter.

\subsection{Compatibility with Quantization}
SVD-based LLM compression methods and quantization are two orthogonal techniques. To demonstrate that our method can be integrated with quantization, we adopt GPTQ~\cite{frantar2022gptq} to quantize LLaMA-2-7B compressed by our method and SVD-LLM. As shown in Table~\ref{quantization}, our method integrates seamlessly with quantization, achieving superior performance compared to SVD-LLM.

\subsection{Efficiency Results}
ERC-SVD not only preserves competitive model performance but also achieves substantial inference speedup on hardware. We evaluate the throughput of compressed models on an NVIDIA A100 GPU, shown in Figure~\ref{throughput_comparison}. Models compressed by ERC-SVD consistently deliver faster inference than the original model. The speedup grows more pronounced as the batch size grows, indicating that ERC-SVD scales more efficiently under larger workloads.

\section{Conclusion}\label{conclusion}
In this work, we propose ERC-SVD, a novel post-training SVD-based compression method for LLMs, formulated from an error-controlled perspective. 
ERC-SVD effectively leverages the residual matrix resulting from SVD truncation to reduce the truncation loss and enhance layer-wise reconstruction accuracy. 
Furthermore, it selectively compresses only the last few layers of the model, thereby significantly mitigating error propagation.
Extensive experiments across various LLM families and benchmark datasets demonstrate that ERC-SVD consistently outperforms existing SVD-based baselines under various settings.

\section*{Acknowledgement}
This paper is supported by Young Scientists Fund of the National Natural Science Foundation of China (NSFC) (No. 62506305), Zhejiang Leading Innovative and Entrepreneur Team Introduction Program (No. 2024R01007), Key Research and Development Program of Zhejiang Province (No. 2025C01026), Scientific Research Project of Westlake University (No. WU2025WF003), Chinese Association for Artificial Intelligence (CAAI) \& Ant Group Research Fund - AGI Track (No. 2025CAAI-ANT-13). It is also supported by the research funds of the National Talent Program and Hangzhou Municipal Talent Program.

% Reference
% For natbib users:

% \clearpage
\phantomsection
\bibliography{CPAL-2025-template/reference}

%%%%%%%%%%%%%%%%%%%%%%%%%%%%%%%%%%%%%%%%%%%%%%%%%%%%%%%%%%%%
\clearpage
\phantomsection
\appendix
\section{Appendix} \label{Appendix}

Here, we provide further details that are not discussed in the main paper and include extra experimental results. The appendix is structured as follows:

\startcontents[appendices]
\printcontents[appendices]{l}{1}{\setcounter{tocdepth}{3}}

\subsection{Pseudocode} \label{pseudpcode}
Algorithm~\ref{alg_rc} and Algorithm~\ref{alg_plc} present the pseudocode for \textbf{residual compensation} and \textbf{partial-layer compression}, respectively.
Algorithm~\ref{alg_plc} identifies the optimal number of last layers $k$ to compress, along with their corresponding layer compression ratio $R_l$.
Specifically, for a model with $N$ layers, we iterate over candidate values of $k'$ using a step size $s$ and compute the associated $R_l'$ for each.
For each candidate configuration, we invoke Algorithm~\ref{alg_rc} to perform the compression, after which we compute the final-layer error relative to the original model.
The configuration yielding the lowest final-layer error is selected as the optimal compression setting.

\begin{algorithm}[ht]
    \caption{Pseudocode of Residual Compensation}
    \label{alg_rc}
    \renewcommand{\algorithmicrequire}{\textbf{Input:}}
    \renewcommand{\algorithmicensure}{\textbf{Output:}}
    
    \begin{algorithmic}[1]
        \REQUIRE Original LLM: $M$, weight matrix: $\boldsymbol{W}_i\in\mathbb{R}^{m\times n}$, layer compression ratio: $R_l$, residual compensation rank: $r_r$   %%input
        \ENSURE Compressed weight matrix set $\mathrm{Set}_{\boldsymbol{W'}}(i)$ of layer $i$     %%output
        
        \STATE $CD \gets$ Randomly collect calibration samples from the dataset

        \STATE $\mathrm{Set}_{\boldsymbol{S}} \gets$ \textsc{Whitening}($M$, $CD$), $\mathrm{Set}_{\boldsymbol{W'}}(i) \gets \emptyset$ \hfill $\triangleright$ Initialize sets of weight matrices

        \STATE $r = (1-R_l)(m \cdot n)/(m+n) $, $r_i = r-r_r$ \hfill $\triangleright$ Calculate the intermediate rank
            
        \STATE $\boldsymbol{S}_i \gets \mathrm{Set}_{\boldsymbol{S}}(i)$ \hfill $\triangleright$ Extract scaling matrices of current weight matrices

        \STATE $\boldsymbol{U}_{i,r_i}$, $\boldsymbol{\Sigma}_{i,r_i}$, $\boldsymbol{V}_{i,r_i}^T$ $\gets$ $\boldsymbol{W}_{i,r_i}\boldsymbol{S}_i^{-1}$ $\gets$ \textsc{SVD\_Trunc}($\boldsymbol{W}_i\boldsymbol{S}_i$)$\boldsymbol{S}_i^{-1}$ \hfill $\triangleright$ SVD truncation on weight matrices

        \STATE $\boldsymbol{R}_{i}$ $\gets$ \textsc{Cal\_Res}($\boldsymbol{W}_i$, $\boldsymbol{W}_{i,r_i}$) \hfill $\triangleright$ Calculate residual matrices

        \STATE $\boldsymbol{U}_{i,r_r}$, $\boldsymbol{\Sigma}_{i,r_r}$, $\boldsymbol{V}_{i,r_r}^T$ $\gets$ $\boldsymbol{R}_{i,r_r}$ $\gets$ \textsc{SVD\_Trunc}($\boldsymbol{R}_i$) \hfill $\triangleright$ SVD truncation on residual matrices

        \STATE $\hat{\boldsymbol{U}}_{i,r_r}$ $\gets$ \textsc{Mul}($\boldsymbol{U}_{i,r_r}$, $\sqrt{\boldsymbol{\Sigma}_{i,r_r}}$), $\hat{\boldsymbol{V}}_{i,r_r}$ $\gets$ \textsc{Mul}($\sqrt{\boldsymbol{\Sigma}_{i,r_r}}$, $\boldsymbol{V}_{i,r_r}^T$) \hfill $\triangleright$ Absorb singular values

        \STATE $\hat{\boldsymbol{U}}_{i,r_i}$ $\gets$ \textsc{Mul}($\boldsymbol{U}_{i,r_i}$, $\sqrt{\boldsymbol{\Sigma}_{i,r_i}}$), $\hat{\boldsymbol{V}}_{i,r_i}$ $\gets$ \textsc{Mul}($\sqrt{\boldsymbol{\Sigma}_{i,r_i}}$, $\boldsymbol{V}_{i,r_i}^T$) 
                
        \STATE $\hat{\boldsymbol{U}}_{i,r}$ $\gets$ \textsc{Comb}($\hat{\boldsymbol{U}}_{i,r_r}$, $\hat{\boldsymbol{U}}_{i,r_i}$), $\hat{\boldsymbol{V}}_{i,r}$ $\gets$ \textsc{Comb}($\hat{\boldsymbol{V}}_{i,r_r}$, $\hat{\boldsymbol{V}}_{i,r_i}$) \hfill $\triangleright$ Combine weight matrices  

        \STATE $\mathrm{Set}_{\boldsymbol{W'}}(i)$ $\gets$ ($\hat{\boldsymbol{U}}_{i,r}$, $\hat{\boldsymbol{V}}_{i,r}$) $\cup$ $\mathrm{Set}_{\boldsymbol{W'}}(i)$ \hfill $\triangleright$ Append decomposed weight matrices

        \RETURN $\mathrm{Set}_{\boldsymbol{W'}}(i)$
    \end{algorithmic}
\end{algorithm}

\begin{algorithm}[ht]
  \caption{Pseudocode of Partial-layer Compression}
    \label{alg_plc}
    \renewcommand{\algorithmicrequire}{\textbf{Input:}}
    \renewcommand{\algorithmicensure}{\textbf{Output:}}
  \begin{algorithmic}[1]
    \REQUIRE Original LLM: $M$, number of model layers: $N$, overall compression ratio: $R_o$, step: $s$ %%input
    \ENSURE Number of layers to compress: $k$, layer compression ratio: $R_l$   %%output

    \STATE $CD \gets$ Randomly collect calibration samples from the dataset
    
    \STATE $\mathrm{Set}_{\boldsymbol{W}} \gets M$ \hfill $\triangleright$ Extract the set of weight matrices in $M$
    
    \STATE $\mathrm{Set}_{k'} \gets \{\,k' \mid k'=s,2s,\dots,N-s\}$ \hfill $\triangleright$ Obtain candidate compressed layer numbers
    \STATE $ Lowest\_Err \gets +\infty$
    \FOR{each $k'$ in $\mathrm{Set}_{k'}$}
        \STATE $R_l' \gets (N \cdot R_o) / k'$ \hfill $\triangleright$ Calculate corresponding layer compression ratio 
        \IF{$R_l' \ge 0$}
            \STATE $M_{TMP} \gets$ \textsc{Deep\_Copy}$(M)$
            \FOR{$i \in [N-k'+1, N]$}
                \STATE $\textbf{W}_i \gets$ $\mathrm{Set}_{\boldsymbol{W}}(i)$ \hfill $\triangleright$ Initialize weight matrices to compress
                \STATE $\mathrm{Set}_{\boldsymbol{W'}}(i) \gets$ \textsc{Residual Compensation}($M$, $\textbf{W}_i$, $R_l'$, $r_r$)
                \STATE Replace weights of layer $i$ in $M_{TMP}$ with $\mathrm{Set}_{\boldsymbol{W'}}(i)$
            \ENDFOR
            \STATE $Err \gets$ \textsc{Cal\_Error}($M, M_{TMP}, CD, Layer(N)$) \hfill $\triangleright$ Calculate the last layer output error
            \IF{$Err < Lowest\_Err$} 
                \STATE $Lowest\_Err \gets Err$ \hfill $\triangleright$ Search for the lowest layer-wise error
                \STATE $k   \gets k'$
                \STATE $R_l \gets R_l'$
            \ENDIF
        \ENDIF
    \ENDFOR
    \RETURN $k, R_l$
  \end{algorithmic}
\end{algorithm}

\subsection{Layer-wise Error Comparison}
\begin{figure}[ht]
  \centering

  \begin{subfigure}[t]{0.49\linewidth}
    \centering
    \includegraphics[width=\linewidth]{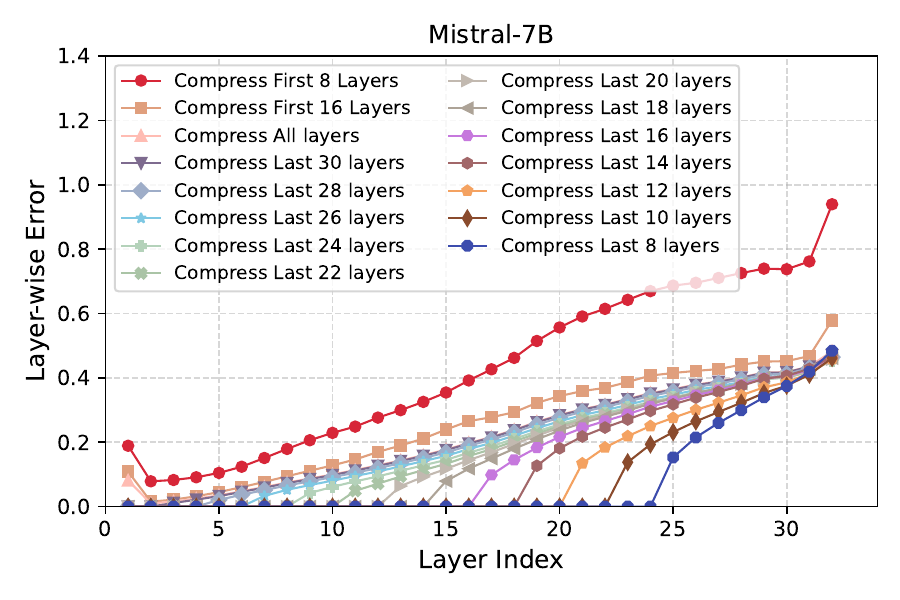}
  \end{subfigure}
  \begin{subfigure}[t]{0.49\linewidth}
    \centering
    \includegraphics[width=\linewidth]{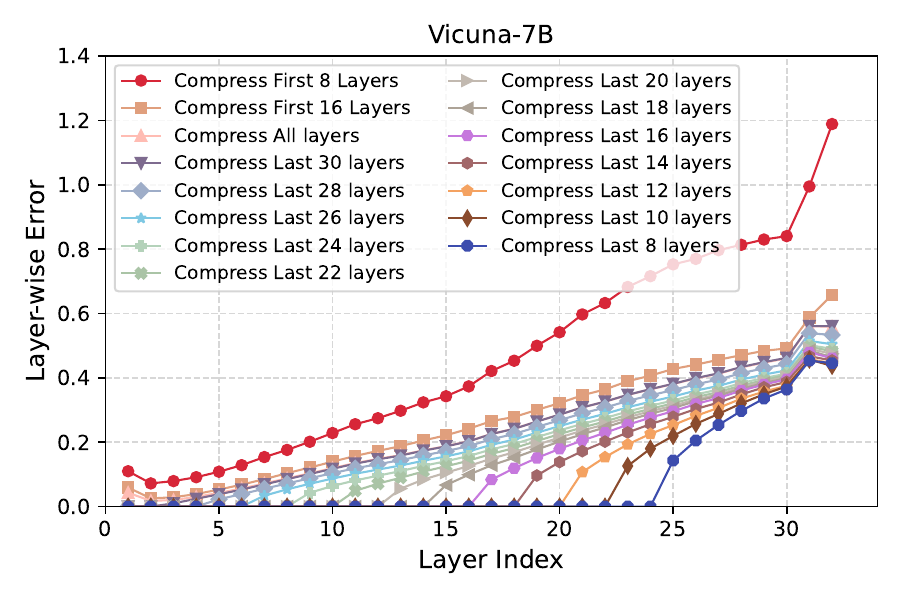}
  \end{subfigure}

  \caption{
    Layer-wise error comparison between the original model, Mistral-7B, and Vicuna-7B compressed by ERC-SVD with different layer selection strategies on WikiText-2. The overall compression ratio is 20\%, and all layer selection strategies strictly adhere to the compression constraint.
  }
  \label{layer-wise_error_comparison_llm-1}
\end{figure}

As discussed in Section~\ref{sec_plc}, the same observation also holds here. We conduct experiments on Mistral-7B~\cite{jiang2023mistral} and Vicuna-7B~\cite{chiang2023vicuna}, and the results are presented in Figure~\ref{layer-wise_error_comparison_llm-1}.

\subsection{Additional Results}
In this section, we present additional results from four perspectives: Section~\ref{more_results_on_large} reports zero-shot accuracy on larger-scale LLMs, Section~\ref{more_results_on_multiple} reports zero-shot accuracy across multiple LLM families, Section~\ref{ablation_study_50_60} presents ablation results under compression ratios 40\%, 50\%, and 60\%, and Section~\ref{calibration-data} demonstrates the impact of calibration data.

\subsubsection{Results on Larger-scale LLMs} \label{more_results_on_large}
In Table~\ref{larger-llm-acc}, we present zero-shot accuracy results for LLaMA-13B~\cite{touvron2023llama}, LLaMA-30B~\cite{touvron2023llama}, LLaMA-2-13B~\cite{touvron2023llama2}, OPT-13B~\cite{zhang2022opt}, and OPT-30B~\cite{zhang2022opt}. Across these evaluation datasets, ERC-SVD consistently outperforms SVD-LLM, with only one minor exception.
\begin{table}[t]
\centering
\caption{Zero-shot accuracy results across larger-scale LLMs under 20\% compression ratio.}
\label{larger-llm-acc}
% \vspace{1mm}
% \vspace{-5pt}
\small
\resizebox{\textwidth}{!}{
\begin{tabular}{c|c|cccccccc}
    \toprule
     \textsc{Model} & \textsc{Method} & Openb.$\uparrow$  & ARC\_e$\uparrow$ & WinoG.$\uparrow$ & HellaS.$\uparrow$ & ARC\_c$\uparrow$ & PIQA$\uparrow$ & MathQA$\uparrow$ & Average\textcolor{green!60!black}{$\uparrow$}\\
    \midrule
        \multirow{2}{*}{LLaMA-13B}
              
                               & \cellcolor{white} SVD-LLM & \cellcolor{white} 0.27 & \cellcolor{white} 0.62 & \cellcolor{white} 0.66 & \cellcolor{white} 0.46 & \cellcolor{white} 0.33 & \cellcolor{white} 0.72 & \cellcolor{white} 0.26 & \cellcolor{white} 0.47 \\
        \cmidrule{2-10}
                               & \cellcolor{cyan!10} \textbf{ERC-SVD} & \cellcolor{cyan!10} \textbf{0.28} & \cellcolor{cyan!10} \textbf{0.69} & \cellcolor{cyan!10} \textbf{0.69} & \cellcolor{cyan!10} \textbf{0.54} & \cellcolor{cyan!10} \textbf{0.39} & \cellcolor{cyan!10} \textbf{0.75} & \cellcolor{cyan!10} \textbf{0.26} & \cellcolor{cyan!10} \textbf{0.52} \\

    \midrule
        \multirow{2}{*}{LLaMA-30B}
              
                              & \cellcolor{white} SVD-LLM & \cellcolor{white} 0.29 & \cellcolor{white} 0.72 & \cellcolor{white} 0.73 & \cellcolor{white} 0.51 & \cellcolor{white} 0.38 & \cellcolor{white} 0.74 & \cellcolor{white} 0.27 & \cellcolor{white} 0.51 \\
        \cmidrule{2-10}
                               & \cellcolor{cyan!10} \textbf{ERC-SVD} & \cellcolor{cyan!10} \textbf{0.30} & \cellcolor{cyan!10} \textbf{0.73} & \cellcolor{cyan!10} \textbf{0.73} & \cellcolor{cyan!10} \textbf{0.59} & \cellcolor{cyan!10} \textbf{0.43} & \cellcolor{cyan!10} \textbf{0.78} & \cellcolor{cyan!10} \textbf{0.29} & \cellcolor{cyan!10} \textbf{0.55} \\
    \midrule
        \multirow{2}{*}{LLaMA-2-13B}
              
                              & \cellcolor{white} SVD-LLM & \cellcolor{white} 0.27 & \cellcolor{white} 0.63 & \cellcolor{white} 0.66 & \cellcolor{white} 0.44 & \cellcolor{white} 0.31 & \cellcolor{white} 0.71 & \cellcolor{white} 0.26 & \cellcolor{white} 0.47 \\
        \cmidrule{2-10}
                               & \cellcolor{cyan!10} \textbf{ERC-SVD} & \cellcolor{cyan!10} \textbf{0.31} & \cellcolor{cyan!10} \textbf{0.67} & \cellcolor{cyan!10} \textbf{0.68} & \cellcolor{cyan!10} \textbf{0.47} & \cellcolor{cyan!10} \textbf{0.38} & \cellcolor{cyan!10} \textbf{0.74} & \cellcolor{cyan!10} \textbf{0.27} & \cellcolor{cyan!10} \textbf{0.50} \\

    \midrule
        \multirow{2}{*}{OPT-13B}
              
                              & \cellcolor{white} SVD-LLM & \cellcolor{white} 0.24 & \cellcolor{white} 0.62 & \cellcolor{white} 0.64 & \cellcolor{white} \textbf{0.47} & \cellcolor{white} 0.30 & \cellcolor{white} 0.72 & \cellcolor{white} 0.24 & \cellcolor{white} 0.46 \\
        \cmidrule{2-10}
                               & \cellcolor{cyan!10} \textbf{ERC-SVD} & \cellcolor{cyan!10} \textbf{0.24} & \cellcolor{cyan!10} \textbf{0.63} & \cellcolor{cyan!10} \textbf{0.65} & \cellcolor{cyan!10} 0.45 & \cellcolor{cyan!10} \textbf{0.30} & \cellcolor{cyan!10} \textbf{0.73} & \cellcolor{cyan!10} \textbf{0.24} & \cellcolor{cyan!10} \textbf{0.46} \\

    \midrule
        \multirow{2}{*}{OPT-30B}
              
                              & \cellcolor{white} SVD-LLM & \cellcolor{white} \textbf{0.29} & \cellcolor{white} 0.67 & \cellcolor{white} 0.65 & \cellcolor{white} \textbf{0.51} & \cellcolor{white} 0.33 & \cellcolor{white} 0.75 & \cellcolor{white} 0.25 & \cellcolor{white} 0.49 \\
        \cmidrule{2-10}
                               & \cellcolor{cyan!10} \textbf{ERC-SVD} & \cellcolor{cyan!10} 0.28 & \cellcolor{cyan!10} \textbf{0.68} & \cellcolor{cyan!10} \textbf{0.67} & \cellcolor{cyan!10} 0.50 & \cellcolor{cyan!10} \textbf{0.34} & \cellcolor{cyan!10} \textbf{0.76} & \cellcolor{cyan!10} \textbf{0.25} & \cellcolor{cyan!10} \textbf{0.49} \\
    \bottomrule
\end{tabular}
}
\end{table}

\subsubsection{Results on Multiple LLM Families} \label{more_results_on_multiple}
We also evaluate the zero-shot accuracy of OPT-6.7B~\cite{zhang2022opt}, Mistral-7B~\cite{jiang2023mistral}, and Vicuna-7B~\cite{chiang2023vicuna}. Results are shown in Table~\ref{multiple-llm-acc}, and our method consistently outperforms SVD-LLM across these diverse LLM architectures.
In addition, we conduct experiments on Qwen-3-8B~\cite{yang2025qwen3}, with perplexity and zero-shot accuracy results reported in Table~\ref{new-model-acc}.

\begin{table}[t]
\centering
\caption{Performance of Qwen-3-8B under 20\% to 60\% compression ratios (``\textsc{Ratio}'').}
\label{new-model-acc}
% \vspace{1mm}
% \vspace{-5pt}
\small
\resizebox{\textwidth}{!}{
\begin{tabular}{c|c|ccc|cccccccc}
    \toprule
     \textsc{Ratio} & \textsc{Method} & WikiText-2\textcolor{blue}{$\downarrow$} & PTB\textcolor{blue}{$\downarrow$} & C4\textcolor{blue}{$\downarrow$} & Openb.$\uparrow$  & ARC\_e$\uparrow$ & WinoG.$\uparrow$ & HellaS.$\uparrow$ & ARC\_c$\uparrow$ & PIQA$\uparrow$ & MathQA$\uparrow$ & Average\textcolor{green!60!black}{$\uparrow$}\\
    \midrule
                               & \cellcolor{white} \textcolor{gray}{Original} & \cellcolor{white} \textcolor{gray}{9.71} & \cellcolor{white} \textcolor{gray}{15.43} & \cellcolor{white} \textcolor{gray}{15.52} & \cellcolor{white} \textcolor{gray}{0.31} & \cellcolor{white} \textcolor{gray}{0.83} & \cellcolor{white} \textcolor{gray}{0.68} & \cellcolor{white} \textcolor{gray}{0.57} & \cellcolor{white} \textcolor{gray}{0.56} & \cellcolor{white} \textcolor{gray}{0.77} & \cellcolor{white} \textcolor{gray}{0.49} & \cellcolor{white} \textcolor{gray}{0.60}\\   
    \midrule
    \multirow{2}{*}{20\%}
                               & \cellcolor{white} SVD-LLM & \cellcolor{white} 37.52 & \cellcolor{white} \textbf{40.73} & \cellcolor{white} 47.25 & \cellcolor{white} \textbf{0.20} & \cellcolor{white} 0.50 & \cellcolor{white} 0.55 & \cellcolor{white} 0.35 & \cellcolor{white} 0.24 & \cellcolor{white} 0.64 & \cellcolor{white} 0.22 & \cellcolor{white} 0.39 \\
                               
                               & \cellcolor{cyan!10} \textbf{ERC-SVD} & \cellcolor{cyan!10} \textbf{35.11} & \cellcolor{cyan!10} 42.77 & \cellcolor{cyan!10} \textbf{43.10} & \cellcolor{cyan!10} 0.19 & \cellcolor{cyan!10} \textbf{0.54} & \cellcolor{cyan!10} \textbf{0.58} & \cellcolor{cyan!10} \textbf{0.35} & \cellcolor{cyan!10} \textbf{0.25} & \cellcolor{cyan!10} \textbf{0.66} & \cellcolor{cyan!10} \textbf{0.24} & \cellcolor{cyan!10} \textbf{0.40} \\
    \midrule
    \multirow{2}{*}{30\%}
                               & \cellcolor{white} SVD-LLM & \cellcolor{white} 72.33 & \cellcolor{white} 67.68 & \cellcolor{white} 86.56 & \cellcolor{white} 0.15 & \cellcolor{white} \textbf{0.42} & \cellcolor{white} 0.53 & \cellcolor{white} 0.30 & \cellcolor{white} 0.19 & \cellcolor{white} 0.60 & \cellcolor{white} 0.21 & \cellcolor{white} 0.34 \\
                               
                               & \cellcolor{cyan!10} \textbf{ERC-SVD} & \cellcolor{cyan!10} \textbf{66.89} & \cellcolor{cyan!10} \textbf{71.48} & \cellcolor{cyan!10} \textbf{82.28} & \cellcolor{cyan!10} \textbf{0.16} & \cellcolor{cyan!10} 0.41 & \cellcolor{cyan!10} \textbf{0.53} & \cellcolor{cyan!10} \textbf{0.30} & \cellcolor{cyan!10} \textbf{0.20} & \cellcolor{cyan!10} \textbf{0.61} & \cellcolor{cyan!10} \textbf{0.22} & \cellcolor{cyan!10} \textbf{0.35} \\
    \midrule
    \multirow{2}{*}{40\%}
                               & \cellcolor{white} SVD-LLM & \cellcolor{white} 143.28 & \cellcolor{white} 122.55 & \cellcolor{white} 162.99 & \cellcolor{white} 0.12 & \cellcolor{white} 0.34 & \cellcolor{white} 0.51 & \cellcolor{white} 0.28 & \cellcolor{white} 0.16 & \cellcolor{white} \textbf{0.58} & \cellcolor{white} 0.22 & \cellcolor{white} 0.32 \\
                               
                               & \cellcolor{cyan!10} \textbf{ERC-SVD} & \cellcolor{cyan!10} \textbf{135.66} & \cellcolor{cyan!10} \textbf{113.34} & \cellcolor{cyan!10} \textbf{154.32} & \cellcolor{cyan!10} \textbf{0.12} & \cellcolor{cyan!10} \textbf{0.35} & \cellcolor{cyan!10} \textbf{0.52} & \cellcolor{cyan!10} \textbf{0.29} & \cellcolor{cyan!10} \textbf{0.17} & \cellcolor{cyan!10} 0.57 & \cellcolor{cyan!10} \textbf{0.22} & \cellcolor{cyan!10} \textbf{0.32} \\
    \midrule
    \multirow{2}{*}{50\%}
                               & \cellcolor{white} SVD-LLM & \cellcolor{white} 274.04 & \cellcolor{white} 226.30 & \cellcolor{white} 308.11 & \cellcolor{white} 0.12 & \cellcolor{white} 0.30 & \cellcolor{white} 0.50 & \cellcolor{white} 0.27 & \cellcolor{white} 0.18 & \cellcolor{white} 0.55 & \cellcolor{white} 0.20 & \cellcolor{white} 0.30 \\
                               
                               & \cellcolor{cyan!10} \textbf{ERC-SVD} & \cellcolor{cyan!10} \textbf{247.57} & \cellcolor{cyan!10} \textbf{170.82} & \cellcolor{cyan!10} \textbf{294.00} & \cellcolor{cyan!10} \textbf{0.12} & \cellcolor{cyan!10} \textbf{0.30} & \cellcolor{cyan!10} \textbf{0.50} & \cellcolor{cyan!10} \textbf{0.28} & \cellcolor{cyan!10} \textbf{0.18} & \cellcolor{cyan!10} \textbf{0.56} & \cellcolor{cyan!10} \textbf{0.22} & \cellcolor{cyan!10} \textbf{0.31} \\
    \midrule
    \multirow{2}{*}{60\%}
                               & \cellcolor{white} SVD-LLM & \cellcolor{white} 462.53 & \cellcolor{white} 338.39 & \cellcolor{white} 528.22 & \cellcolor{white} 0.11 & \cellcolor{white} 0.27 & \cellcolor{white} \textbf{0.48} & \cellcolor{white} 0.26 & \cellcolor{white} 0.18 & \cellcolor{white} \textbf{0.54} & \cellcolor{white} 0.20 & \cellcolor{white} 0.29 \\
                               
                               & \cellcolor{cyan!10} \textbf{ERC-SVD} & \cellcolor{cyan!10} \textbf{460.73} & \cellcolor{cyan!10} \textbf{287.19} & \cellcolor{cyan!10} \textbf{522.29} & \cellcolor{cyan!10} \textbf{0.11} & \cellcolor{cyan!10} \textbf{0.27} & \cellcolor{cyan!10} 0.46 & \cellcolor{cyan!10} \textbf{0.27} & \cellcolor{cyan!10} \textbf{0.19} & \cellcolor{cyan!10} 0.53 & \cellcolor{cyan!10} \textbf{0.20} & \cellcolor{cyan!10} \textbf{0.29} \\

    \bottomrule
\end{tabular}
}
\end{table}

\subsubsection{Additional Ablation Results} \label{ablation_study_50_60}
The ablation results for LLaMA-2-7B~\cite{touvron2023llama2} under 40\%, 50\%, and 60\% compression ratios are presented in Table~\ref{more_ablation_study}.
A similar trend can also be observed here: incorporating both REC and PLC leads to a substantial reduction in perplexity across all settings.

\begin{table}[ht]
\centering
\caption{Ablation results of residual compensation and partial-layer compression on LLaMA-2-7B~\cite{touvron2023llama2} under 40\%, 50\%, and 60\% compression ratios.}
\label{more_ablation_study}
% \vspace{1mm}
% \vspace{-5pt}
\small
\resizebox{\textwidth}{!}{
\begin{tabular}{c|c|cc|c|cccccccc}
    \toprule
     \textsc{Ratio} & \textsc{Method} & REC & PLC & C4\textcolor{blue}{$\downarrow$} & Openb.$\uparrow$  & ARC\_e$\uparrow$ & WinoG.$\uparrow$ & HellaS.$\uparrow$ & ARC\_c$\uparrow$ & PIQA$\uparrow$ & MathQA$\uparrow$ & Average\textcolor{green!60!black}{$\uparrow$}\\

     \midrule
        \multirow{5}{*}{40\%}  
                               & ASVD & \cellcolor{white} - & \cellcolor{white} - & \cellcolor{white} NaN & \cellcolor{white} 0.15 & \cellcolor{white} 0.25 & \cellcolor{white} 0.50 & \cellcolor{white} 0.26 & \cellcolor{white} 0.22 & \cellcolor{white} 0.52 & \cellcolor{white} 0.18 & \cellcolor{white} 0.30\\
        
                               & SVD-LLM & \cellcolor{white} - & \cellcolor{white} - & \cellcolor{white} 61.96 & \cellcolor{white} 0.16 & \cellcolor{white} 0.35 & \cellcolor{white} 0.55 & \cellcolor{white} 0.30 & \cellcolor{white} 0.20 & \cellcolor{white} 0.57 & \cellcolor{white} 0.23 & \cellcolor{white} 0.34\\

        \cmidrule{2-13}
        & \multirow{3}{*}{\textbf{ERC-SVD}}
                               & \ding{51} & \ding{55} & 54.19 & 0.17 & 0.37 & 0.54 & 0.32 & 0.21 & 0.57 & 0.23 & 0.34\\

                               & & \ding{55} & \ding{51} & 45.13 & 0.18 & 0.40 & 0.55 & 0.33 & 0.24 & 0.61 & 0.21 & 0.36\\

                               & & \cellcolor{cyan!10} \ding{51} & \cellcolor{cyan!10} \ding{51} & \cellcolor{cyan!10} \textbf{43.19}  & \cellcolor{cyan!10} 0.20 & \cellcolor{cyan!10} 0.43 & \cellcolor{cyan!10} 0.57 & \cellcolor{cyan!10} 0.35 & \cellcolor{cyan!10} 0.24 & \cellcolor{cyan!10} 0.63 & \cellcolor{cyan!10} 0.23 & \cellcolor{cyan!10} \textbf{0.38} \\
                               
    \midrule
        \multirow{5}{*}{50\%}  
                               & ASVD & \cellcolor{white} - & \cellcolor{white} - & \cellcolor{white} NaN & \cellcolor{white} 0.13 & \cellcolor{white} 0.26 & \cellcolor{white} 0.50 & \cellcolor{white} 0.25 & \cellcolor{white} 0.23 & \cellcolor{white} 0.50 & \cellcolor{white} 0.20 & \cellcolor{white} 0.30\\
              
                               & SVD-LLM & \cellcolor{white} - & \cellcolor{white} - & \cellcolor{white} 129.71 & \cellcolor{white} 0.14 & \cellcolor{white} 0.30 & \cellcolor{white} 0.50 & \cellcolor{white} 0.28 & \cellcolor{white} 0.20 & \cellcolor{white} 0.54 & \cellcolor{white} 0.23 & \cellcolor{white} 0.31\\
        \cmidrule{2-13}
        & \multirow{3}{*}{\textbf{ERC-SVD}}
                               & \ding{51} & \ding{55} & 126.61 & 0.14 & 0.30 & 0.53 & 0.29 & 0.21 & 0.55 & 0.24 & 0.32\\

                               &  & \ding{55} & \ding{51} & 117.79 & 0.13 & 0.33 & 0.54 & 0.29 & 0.22 & 0.58 & 0.22 & 0.33\\
                                
                               &  & \cellcolor{cyan!10} \ding{51} & \cellcolor{cyan!10} \ding{51} & \cellcolor{cyan!10} \textbf{100.34} & \cellcolor{cyan!10} 0.14 & \cellcolor{cyan!10} 0.35 & \cellcolor{cyan!10} 0.55 & \cellcolor{cyan!10} 0.31 & \cellcolor{cyan!10} 0.22 & \cellcolor{cyan!10} 0.59 & \cellcolor{cyan!10} 0.22 & \cellcolor{cyan!10} \textbf{0.34}\\

    \midrule
        \multirow{5}{*}{60\%}  
                               & ASVD & \cellcolor{white} - & \cellcolor{white} - & \cellcolor{white} NaN & \cellcolor{white} 0.15 & \cellcolor{white} 0.25 & \cellcolor{white} 0.50 & \cellcolor{white} 0.25 & \cellcolor{white} 0.23 & \cellcolor{white} 0.52 & \cellcolor{white} 0.12 & \cellcolor{white} 0.29\\
              
                               & SVD-LLM & \cellcolor{white} - & \cellcolor{white} - & \cellcolor{white} 263.02 & \cellcolor{white} 0.14 & \cellcolor{white} 0.26 & \cellcolor{white} 0.50 & \cellcolor{white} 0.27 & \cellcolor{white} 0.20 & \cellcolor{white} 0.53 & \cellcolor{white} 0.21 & \cellcolor{white} 0.30\\

        \cmidrule{2-13}
        & \multirow{3}{*}{\textbf{ERC-SVD}}
                               & \ding{51} & \ding{55} & 256.38 & 0.13 & 0.26 & 0.49 & 0.27 & 0.20 & 0.53 & 0.23 & 0.30 \\

                               &  & \ding{55} & \ding{51} & 260.01 & 0.14 & 0.29 & 0.50 & 0.26 & 0.18 & 0.53 & 0.21 & 0.30 \\
                                
                               &  & \cellcolor{cyan!10} \ding{51} & \cellcolor{cyan!10} \ding{51} & \cellcolor{cyan!10} \textbf{255.70} & \cellcolor{cyan!10} 0.13 & \cellcolor{cyan!10} 0.29 & \cellcolor{cyan!10} 0.52 & \cellcolor{cyan!10} 0.28 & \cellcolor{cyan!10} 0.21 & \cellcolor{cyan!10} 0.55 & \cellcolor{cyan!10} 0.22 & \cellcolor{cyan!10} \textbf{0.31}\\
    \bottomrule
\end{tabular}
}
\end{table}

\begin{figure}[t]
  \centering
  \begin{minipage}[t]{0.61\textwidth}
    \vspace{0pt}
    \captionof{table}{Zero-shot accuracy (\textcolor{green!60!black}{$\uparrow$}) for seven common sense reasoning datasets on OPT-6.7B, Mistral-7B, and Vicuna-7B under 30\% compression ratio.}
    % \vspace{1.2em}
    \vspace{10pt}
    \label{multiple-llm-acc}
    \small
    \resizebox{\linewidth}{!}{
    \begin{tabular}{c|c|cccccccc}
    \toprule
     \textsc{Model} & \textsc{Method} & Openb.$\uparrow$  & ARC\_e$\uparrow$ & WinoG.$\uparrow$ & HellaS.$\uparrow$ & ARC\_c$\uparrow$ & PIQA$\uparrow$ & MathQA$\uparrow$ & Average\textcolor{green!60!black}{$\uparrow$}\\
    \midrule
        \multirow{2}{*}{OPT-6.7B}
              
                               & \cellcolor{white} SVD-LLM & \cellcolor{white} \textbf{0.27} & \cellcolor{white} 0.40 & \cellcolor{white} 0.49 & \cellcolor{white} 0.33 & \cellcolor{white} 0.22 & \cellcolor{white} 0.50 & \cellcolor{white} 0.20 & \cellcolor{white} 0.34\\
        \cmidrule{2-10}
                               & \cellcolor{cyan!10} \textbf{ERC-SVD} & \cellcolor{cyan!10} 0.22 & \cellcolor{cyan!10} \textbf{0.56} & \cellcolor{cyan!10} \textbf{0.61} & \cellcolor{cyan!10} \textbf{0.40} & \cellcolor{cyan!10} \textbf{0.25} & \cellcolor{cyan!10} \textbf{0.69} & \cellcolor{cyan!10} \textbf{0.24} & \cellcolor{cyan!10} \textbf{0.42}\\

    \midrule
        \multirow{2}{*}{Mistral-7B}
              
                               & \cellcolor{white} SVD-LLM & \cellcolor{white} 0.13 & \cellcolor{white} 0.44 & \cellcolor{white} 0.53 & \cellcolor{white} 0.30 & \cellcolor{white} 0.20 & \cellcolor{white} 0.62 & \cellcolor{white} 0.20 & \cellcolor{white} 0.34\\
        \cmidrule{2-10}
                               & \cellcolor{cyan!10} \textbf{ERC-SVD} & \cellcolor{cyan!10} \textbf{0.14} & \cellcolor{cyan!10} \textbf{0.46} & \cellcolor{cyan!10} \textbf{0.58} & \cellcolor{cyan!10} \textbf{0.33} & \cellcolor{cyan!10} \textbf{0.22} & \cellcolor{cyan!10} \textbf{0.63} & \cellcolor{cyan!10} \textbf{0.24} & \cellcolor{cyan!10} \textbf{0.37}\\
    \midrule
        \multirow{2}{*}{Vicuna-7B}
              
                               & \cellcolor{white} SVD-LLM & \cellcolor{white} 0.22 & \cellcolor{white} 0.50 & \cellcolor{white} 0.56 & \cellcolor{white} 0.35 & \cellcolor{white} 0.26 & \cellcolor{white} 0.63 & \cellcolor{white} 0.21 & \cellcolor{white} 0.39\\
        \cmidrule{2-10}
                               & \cellcolor{cyan!10} \textbf{ERC-SVD} & \cellcolor{cyan!10} \textbf{0.23} & \cellcolor{cyan!10} \textbf{0.51} & \cellcolor{cyan!10} \textbf{0.60} & \cellcolor{cyan!10} \textbf{0.39} & \cellcolor{cyan!10} \textbf{0.30} & \cellcolor{cyan!10} \textbf{0.67} & \cellcolor{cyan!10} \textbf{0.22} & \cellcolor{cyan!10} \textbf{0.42}\\
    \bottomrule
\end{tabular}
    }
  \end{minipage}%
  \hfill
  \begin{minipage}[t]{0.34\textwidth}
    % \vspace{0pt}
    \centering
    \captionof{table}{Results on Wikitext-2 under different compression ratios (``\textsc{Ratio}'') of LLaMA-2-7B.}
    \vspace{1pt}
    \label{pruning_comparision}
    \small
    \resizebox{\linewidth}{!}{
    \begin{tabular}{ccc}
        \toprule
            \textsc{Ratio} & LLM-Pruner & \textbf{ERC-SVD} \\
        \midrule
            20\% & 8.09  & 7.63  \\
            30\% & 12.59 & 10.32 \\
            40\% & 20.36 & 14.17 \\
            50\% & 40.97 & 24.26 \\
            60\% & 114.23 & 58.88 \\
\bottomrule
\end{tabular}
}
  \end{minipage}
\end{figure}

\subsubsection{Impact of calibration data} \label{calibration-data}
\begin{figure}[t]

  \begin{minipage}[t]{0.38\textwidth}
    \vspace{0pt}
    \centering
    \includegraphics[width=\linewidth]{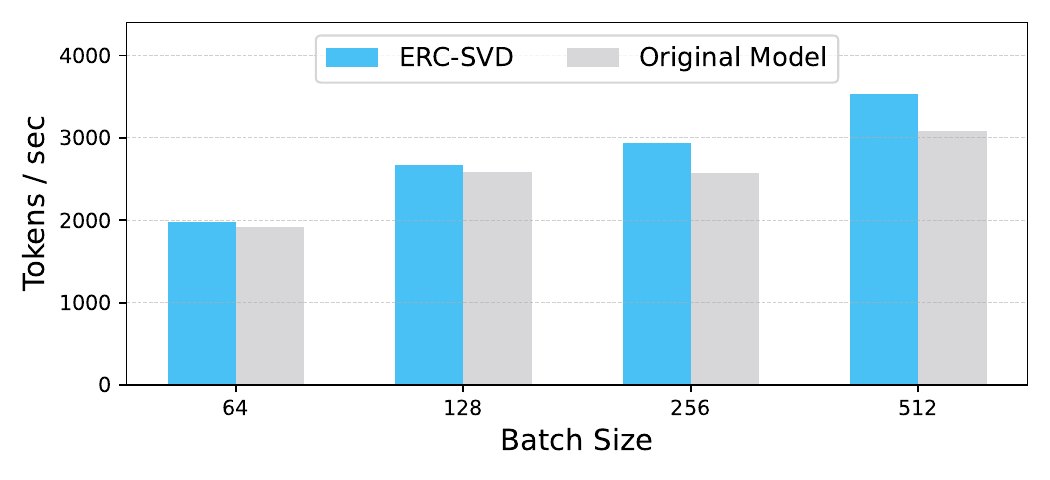}
    \captionof{figure}{Throughput of LLaMA-7B and its 40\% compressed versions. The sequence length is 32.}
    \label{throughput_comparison}
  \end{minipage}%
  \hfill
  \begin{minipage}[t]{0.6\textwidth}
    \vspace{0pt}
    \centering

    \begin{subfigure}[t]{0.48\textwidth}
      \centering
      \includegraphics[width=\linewidth]{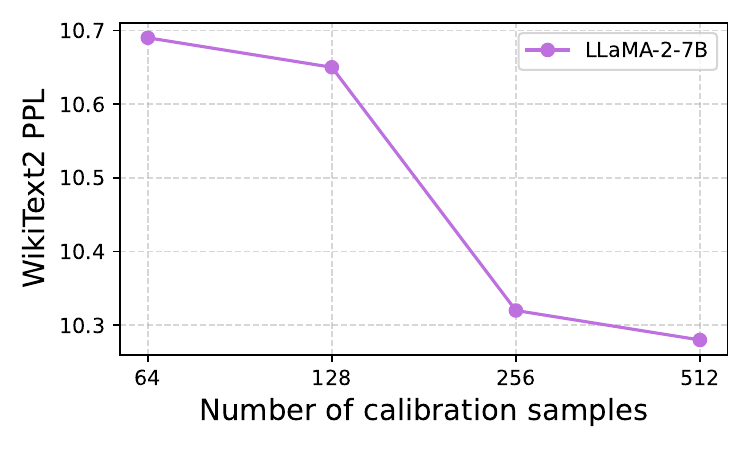}
    \end{subfigure}
    \hfill
    \begin{subfigure}[t]{0.48\textwidth}
      \centering
      \includegraphics[width=\linewidth]{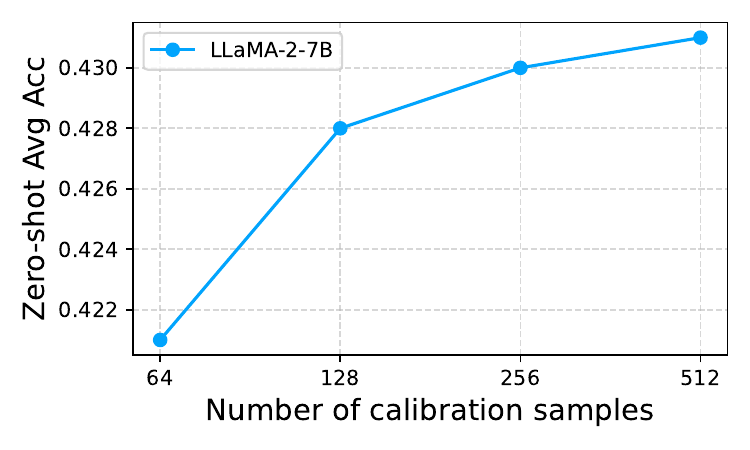}
    \end{subfigure}

    \captionof{figure}{Impact of the number of calibration data samples on LLaMA-2-7B under 30\% compression ratio. (Left) Perplexity (\textcolor{blue}{$\downarrow$}). (Right) Average accuracy (\textcolor{green!60!black}{$\uparrow$}).}
    \label{calibration_samples_summary}
  \end{minipage}

\end{figure}

We study the impact of calibration data from two aspects: the number of calibration samples and the choice of calibration datasets.
Figure~\ref{calibration_samples_summary} shows the evaluation results under 30\% compression ratio.
Performance improves slightly with more calibration samples, though the gains remain modest, indicating that ERC-SVD is robust even with limited calibration data.
Moreover, Table~\ref{different_calibration_dataset} reports results for different calibration datasets. We also evaluate a mixed dataset, constructed by equally combining the three datasets, to assess its effect on model performance.

\begin{figure}[t]
  \centering
  \begin{minipage}[t]{0.4\textwidth}
    \vspace{0pt}
    \captionof{table}{Perplexity (\textcolor{blue}{$\downarrow$}) and zero-shot average accuracy (\textcolor{green!60!black}{$\uparrow$}) of LLaMA-2-7B 20\% compressed by ERC-SVD with different calibration datasets.}
    % \vspace{1.2em}
    \vspace{-5pt}
    \label{different_calibration_dataset}
    \small
    \resizebox{\linewidth}{!}{
    \begin{tabular}{c|ccc|c}
      \toprule
      % \makecell{\textsc{Calibration}\\\textsc{Dataset}} 
                      \textsc{Cali. Data} & WikiText-2\textcolor{blue}{$\downarrow$} & PTB\textcolor{blue}{$\downarrow$} & C4\textcolor{blue}{$\downarrow$} & Avg.\textcolor{green!60!black}{$\uparrow$} \\
      \midrule 
        WikiText-2 & \textbf{7.63} & 45.37 & 14.73 & \textbf{0.48} \\
        PTB        & 9.71 & \textbf{29.43} & 13.29 & 0.46 \\
        C4         & 9.44 & 37.67 & \textbf{11.49} & 0.47 \\
        Mix        & 10.13 & 36.53 & 13.13 & 0.45 \\
      \bottomrule
    \end{tabular}
    }
  \end{minipage}%
  \hfill
  \begin{minipage}[t]{0.55\textwidth}
    \vspace{0pt}
    \centering
    \captionof{table}{Perplexity (\textcolor{blue}{$\downarrow$}) of LLaMA-2-7B compressed by ERC-SVD and SVD-LLM, followed by quantization with GPTQ-8bit~\cite{frantar2022gptq}. Blue arrows within parentheses highlight the relative improvement.}
    \vspace{3.5pt}
    \label{quantization}
    \small
    \resizebox{\linewidth}{!}{
    \begin{tabular}{ccc|ccc}
    \toprule
     \textsc{Ratio} & \textsc{Method} & \textsc{Quantization} & WikiText-2\textcolor{blue}{$\downarrow$} & PTB\textcolor{blue}{$\downarrow$} & C4\textcolor{blue}{$\downarrow$}\\
    \midrule
                            \textcolor{gray}{-}  & \cellcolor{white} \textcolor{gray}{Original} & \cellcolor{white} \textcolor{gray}{GPTQ-8bit} & \cellcolor{white} \textcolor{gray}{5.47} & \cellcolor{white} \textcolor{gray}{26.79} & \cellcolor{white} \textcolor{gray}{7.28} \\
    \midrule
    \multirow{2}{*}{30\%}      & \cellcolor{white} SVD-LLM & \cellcolor{white} GPTQ-8bit & \cellcolor{white} 10.67 & \cellcolor{white} 336.30 & \cellcolor{white} 34.93 \\
                               & \cellcolor{cyan!10} \textbf{ERC-SVD} & \cellcolor{cyan!10} GPTQ-8bit & \cellcolor{cyan!10} \textbf{10.34} (\textcolor{blue}{$\downarrow$3\%}) & \cellcolor{cyan!10} \textbf{79.05} (\textcolor{blue}{$\downarrow$76\%}) & \cellcolor{cyan!10} \textbf{25.02} (\textcolor{blue}{$\downarrow$28\%}) \\
    \bottomrule
\end{tabular}
}
  \end{minipage}
\end{figure}

\subsection{Comparison with Smaller-Scale LLM} \label{gemma}
To further assess the effectiveness of ERC-SVD under relatively high compression ratios, we conduct a comparative evaluation of 50\% and 60\% compressed LLaMA-2-7B against Gemma-2-2B~\cite{gemma_2024}. Table~\ref{compare_gemma} presents the performance comparison between the compressed LLaMA-2-7B models and Gemma-2-2B. Under 50\% compression ratio (approximately 3.5B parameters), LLaMA-2-7B consistently outperforms Gemma-2-2B across most benchmarks. 
Even under a more aggressive 60\% compression ratio (around 2.8B parameters), the compressed model maintains competitive performance, remaining comparable to Gemma-2-2B.

\begin{table}[t]
\centering
\caption{Performance comparison of Gemma-2-2B and LLaMA-2-7B compressed by ERC-SVD under 50\% and 60\% compression ratios (``\textsc{Ratio}'').}
\label{compare_gemma}
% \vspace{1mm}
% \vspace{-5pt}
\small
\resizebox{\textwidth}{!}{
\begin{tabular}{c|cc|ccc|cccccccc}
    \toprule
     \textsc{Model} & \textsc{Ratio} & \textbf{\#Params} & WikiText-2\textcolor{blue}{$\downarrow$} & PTB\textcolor{blue}{$\downarrow$} & C4\textcolor{blue}{$\downarrow$} & Openb.$\uparrow$  & ARC\_e$\uparrow$ & WinoG.$\uparrow$ & HellaS.$\uparrow$ & ARC\_c$\uparrow$ & PIQA$\uparrow$ & MathQA$\uparrow$ & Average\textcolor{green!60!black}{$\uparrow$}\\   
    \midrule
    \multirow{1}{*}{Gemma-2-2B}
                               & \cellcolor{white} - & \cellcolor{white} 3B & \cellcolor{white} 67.78 & \cellcolor{white} 374.60 & \cellcolor{white} 70.78 & \cellcolor{white} 0.13 & \cellcolor{white} 0.39 & \cellcolor{white} 0.51 & \cellcolor{white} 0.31 & \cellcolor{white} 0.19 & \cellcolor{white} 0.60 & \cellcolor{white} 0.23 & \cellcolor{white} 0.33 \\
    \midrule
    \multirow{2}{*}{LLaMA-2-7B}
                               & \cellcolor{cyan!10} 50\% & \cellcolor{cyan!10} 3.5B & \cellcolor{cyan!10} 24.26 & \cellcolor{cyan!10} 286.24 & \cellcolor{cyan!10} 100.34 & \cellcolor{cyan!10} 0.14 & \cellcolor{cyan!10} 0.35 & \cellcolor{cyan!10} 0.55 & \cellcolor{cyan!10} 0.31 & \cellcolor{cyan!10} 0.22 & \cellcolor{cyan!10} 0.59 & \cellcolor{cyan!10} 0.22 & \cellcolor{cyan!10} 0.34 \\
    \cmidrule{2-14}
                               & \cellcolor{cyan!10} 60\% & \cellcolor{cyan!10} 2.8B & \cellcolor{cyan!10} 68.59 & \cellcolor{cyan!10} 991.48 & \cellcolor{cyan!10} 255.70 & \cellcolor{cyan!10} 0.13 & \cellcolor{cyan!10} 0.29 & \cellcolor{cyan!10} 0.52 & \cellcolor{cyan!10} 0.28 & \cellcolor{cyan!10} 0.21 & \cellcolor{cyan!10} 0.55 & \cellcolor{cyan!10} 0.22 & \cellcolor{cyan!10} 0.31 \\
    \bottomrule
\end{tabular}
}
\end{table}
\vspace{-5pt}
\begin{table}[ht]
\centering
\caption{Performance (measured by accuracy (\textcolor{green!60!black}{$\uparrow$})) of original LLaVA-1.5-7B, and its 20\% compressed versions by SVD-LLM and ERC-SVD on VLM benchmarks. The best results are marked in \textbf{bold}.}
% \vspace{-5pt}
\label{some_benchmark}
\small
\resizebox{\textwidth}{!}{
\begin{tabular}{cc|ccccc}
    \toprule
     \textsc{Compression Ratio} & \textsc{Method} & POPE-random\textcolor{green!60!black}{$\uparrow$} & POPE-popular\textcolor{green!60!black}{$\uparrow$} & POPE-adversial\textcolor{green!60!black}{$\uparrow$} & TextVQA\textcolor{green!60!black}{$\uparrow$} & ScienceQA\textcolor{green!60!black}{$\uparrow$}\\
    \midrule
                            \textcolor{gray}{-}  & \cellcolor{white} \textcolor{gray}{Original} & \cellcolor{white} \textcolor{gray}{88.2} & \cellcolor{white} \textcolor{gray}{87.3} & \cellcolor{white} \textcolor{gray}{85.1} & \cellcolor{white} \textcolor{gray}{58.17} & \cellcolor{white} \textcolor{gray}{70.15}\\
    \midrule
    \multirow{2}{*}{20\%}      & \cellcolor{white} SVD-LLM & \cellcolor{white} 82.5 & \cellcolor{white} 83.2 & \cellcolor{white} 77.8 & \cellcolor{white} 30.68 & \cellcolor{white} 49.54\\
                               & \cellcolor{cyan!10} \textbf{ERC-SVD} & \cellcolor{cyan!10} \textbf{90.2} (\textcolor{green!60!black}{$\uparrow$9\%}) & \cellcolor{cyan!10} \textbf{87.7} (\textcolor{green!60!black}{$\uparrow$5\%}) & \cellcolor{cyan!10} \textbf{83.1} (\textcolor{green!60!black}{$\uparrow$7\%}) & \cellcolor{cyan!10} \textbf{50.86} (\textcolor{green!60!black}{$\uparrow$66\%}) & \cellcolor{cyan!10} \textbf{69.54} (\textcolor{green!60!black}{$\uparrow$40\%})\\
    \bottomrule
\end{tabular}
}
\end{table}

\subsection{Comparison with Pruning} \label{pruning}
Table~\ref{pruning_comparision} shows the performance of LLaMA-2-7B compressed by LLM-Pruner~\cite{ma2023llm} and ERC-SVD under different compression ratios on the WikiText-2 dataset. It can be observed that ERC-SVD consistently outperforms the pruning method, achieving a perplexity of 58.88 under 60\% compression ratio, compared to 114.23 for LLM-Pruner.

\subsection{Computation Complexity} \label{inference_efficiency_analysis}
ERC-SVD decomposes the original weight matrix $\boldsymbol{W}\in\mathbb{R}^{m\times n}$ into two low-rank matrices: $\hat{\boldsymbol{U}}_r\in\mathbb{R}^{m\times r}$ and $\hat{\boldsymbol{V}}_r\in\mathbb{R}^{r\times n}$. 
The layer compression ratio $R_l$ is computed as $R_l = 1 - \frac{(m+n)r}{mn}$. 
Under a fixed overall compression ratio $R_o$, the relationship between $R_l$ and $R_o$ is given by $R_l = \frac{N R_o}{k}$, where $N$ is the total number of layers and $k$ denotes the number of last layers to be compressed.

Given an input activation $\boldsymbol{X} \in \mathbb{R}^{n \times m}$, the original output is computed as $\boldsymbol{Y} = \boldsymbol{W} \boldsymbol{X}$. 
In the compressed model layer, an intermediate state is first computed as $\boldsymbol{I} = \hat{\boldsymbol{V}}_r \boldsymbol{X}$, followed by $\boldsymbol{Y} = \hat{\boldsymbol{U}}_r \boldsymbol{I}$.
The computational complexity of the original model is $N \cdot \mathcal{O}(m^2 n)$.
For the compressed model, the first $(N - k)$ layers remain uncompressed and retain a complexity of $(N - k) \cdot \mathcal{O}(m^2 n)$, while the compressed $k$ layers incur a cost of $k \cdot \mathcal{O}(m^2 r + r n m)$.
Thus, the total complexity becomes:
\begin{align}
    (N-k)\cdot\mathcal{O}(m^2n)+k\cdot\mathcal{O}(m^2r+rnm).
    \label{complexity}
\end{align}
And the rank $r$ is given by:
\begin{align}
    r = \frac{mn(1-R_l)}{m+n} = \frac{mn(k-NR_o)}{k(m+n)}.
    \label{rank}
\end{align}
Substituting the expression for $r$ into Equation~\ref{complexity}, we obtain the simplified total complexity:
\begin{align}
    \boxed{N(1-R_o)\cdot\mathcal{O}(m^2n)}
\end{align}
Compared to the original computation complexity $N \cdot \mathcal{O}(m^2 n)$. This indicates that the total computation cost is reduced proportionally to the overall compression ratio.
For example, if $R_o = 40$\%, the compressed model requires only 60\% of the original computational cost.

\subsection{Demonstration of Generated Contents} \label{generated_content}
Table~\ref{tab:zero_shot_text} shows the generation contents of models compressed by ERC-SVD compared to the original model under zero-shot conditions. The results indicate that across various input questions, compressed models consistently produce fluent, coherent, and highly relevant responses. Even at 40\% compression ratio, the compressed model can also deliver accurate and relevant answers to common sense questions. These results highlight the effectiveness of ERC-SVD.

\begin{table}[ht]
\centering
\caption{Generated contents of LLaMA-7B~\cite{touvron2023llama} compressed by ERC-SVD.}
\vspace{1mm}
\label{tab:zero_shot_text}
\renewcommand{\arraystretch}{1.3}
\setlength{\tabcolsep}{6pt}
\small

\resizebox{0.84\textwidth}{!}{
% \begin{tabular}{@{} >{\raggedright\arraybackslash}p{5cm} | c | >{\raggedright\arraybackslash}p{8cm} @{}}
\begin{tabular}{p{5cm}|c|p{8cm}}
\toprule
\textsc{Question} & \makecell{\textsc{Comp.}\\\textsc{Ratio}} & \textsc{Answer} \\
\midrule

\multirow{13}{=}{What is the responsibility of the AI assistant?}
& -   & The AI assistant is responsible for understanding the context and taking appropriate action. The AI assistant is responsible for understanding the intent, context, environment and taking appropriate actions. The AI assistant is responsible for interpreting. \\
\cmidrule{2-3}
& 20\% & The AI assistant should be responsible for providing an answer to questions. This means that when asked a question, you should do your best to answer it and not give any information other than what is requested. You should also consider the tone and intent. \\
\cmidrule{2-3}
& 30\% & The AI Assistant's job is to provide assistance to those who wish for assistance in their lives. It has no duty or obligation to assist those who do not desire assistance, nor does it have any duty to prevent. \\
\cmidrule{2-3}
& 40\% & The AI assistant does not have a specific role. The AI assistant will provide instructions to follow a path, and will give directions as an instruction to follow the path. \\
\midrule

\multirow{14}{=}{What is the universe? Explain this concept.}
& -   & The Universe is everything that exists. It includes all space, time, matter and energy. It also includes everything which has happened or will happen. The universe began about 13.7 billion years ago (the Big Bang) and will. \\
\cmidrule{2-3}
& 20\% & The universe is the whole of space and time. It consists of all things in space and all things that happen, or have happened. There are many theories on how it began; there were theories that it had no beginning at all. \\
\cmidrule{2-3}
& 30\% & The universe consists of everything in it, including matter and objects, space, the cosmos or outer space, and the galaxy system, as well as other galaxies that make up the Universe. The universe has its own life cycle. \\
\cmidrule{2-3}
& 40\% & In the universe, life is very diverse and complex. The universe is made up of many different objects. These objects are divided into groups. Each group consists of different objects that have specific properties. This group contains the universe, which is large. \\
\midrule

\multirow{11}{=}{Why do cars need fuel to run?}
& -   & Fuel is the lifeblood of a car. Without it, there's nothing to power
  the pistons and no way to drive. A car needs a constant supply of fuel. \\
\cmidrule{2-3}
& 20\% &  The fuel in the tank is burned to produce heat, which produces steam that drives a turbine, which turns a crankshaft. The crankshaft turns the wheels of the car, propelling it forward. \\
\cmidrule{2-3}
& 30\% & The fuel is the fuel in a car and it burns and goes up and up. It's important because it burns the gasoline, which makes it go faster. It's important because you can get energy out. \\
\cmidrule{2-3}
& 40\% & The fuel gives energy for the cars. The gasoline is used by a motor, which produces heat and air, heating up the engine. This allows the engines to travel faster, allowing them to reach higher speed. \\
\midrule

\multirow{9}{=}{Why do people feel hungry after exercise?}
& -   &  The answer is because the body burns more calories when you are exercising. You may not realize that the body uses up energy to digest food and absorb nutrients from it.  \\
\cmidrule{2-3}
& 20\% &  Because the body is more efficient at converting food into energy when inactive, so when active it needs to convert food into energy faster. This is why people often get hungry after exercise. \\
\cmidrule{2-3}
& 30\% & The reason people feel hungry after an activity is because they burn energy and energy is metabolised in the body.  \\
\cmidrule{2-3}
& 40\% & The reason the body is hungry after a long athletic activity was discovered by scientists in 1976. \\
\bottomrule

\end{tabular}
}
\end{table}

\subsection{ERC-SVD on Vision-Language Model}\label{ressvd:vlm}
In this section, we apply ERC-SVD to the language part of vision-language models (VLMs) such as LLaVA~\cite{liu2024improved}, and evaluate both quantitative and qualitative performance.

\subsubsection{Quantitative Results} \label{sec:benchmark}
We compare the performance of LLaVA-1.5-7B~\cite{liu2024improved} compressed using SVD-LLM~\cite{wang2024svd} and ERC-SVD under 20\% compression ratio. 
We report results on several benchmarks: POPE~\cite{li2023evaluating}, TextVQA~\cite{singh2019towards}, and ScienceQA~\cite{lu2022learn}.
The results, shown in Table~\ref{some_benchmark}, indicate that ERC-SVD consistently outperforms SVD-LLM across all benchmarks. 
Notably, it achieves substantial relative improvements, 66\% on TextVQA and 40\% on ScienceQA.
Moreover, on the POPE-random and POPE-popular subsets, the model compressed by ERC-SVD even surpasses the original LLaVA-1.5-7B.

\subsubsection{Qualitative Results} \label{sec:qualitative}
\begin{figure}[h]
  \centering
  \includegraphics[width=\linewidth]{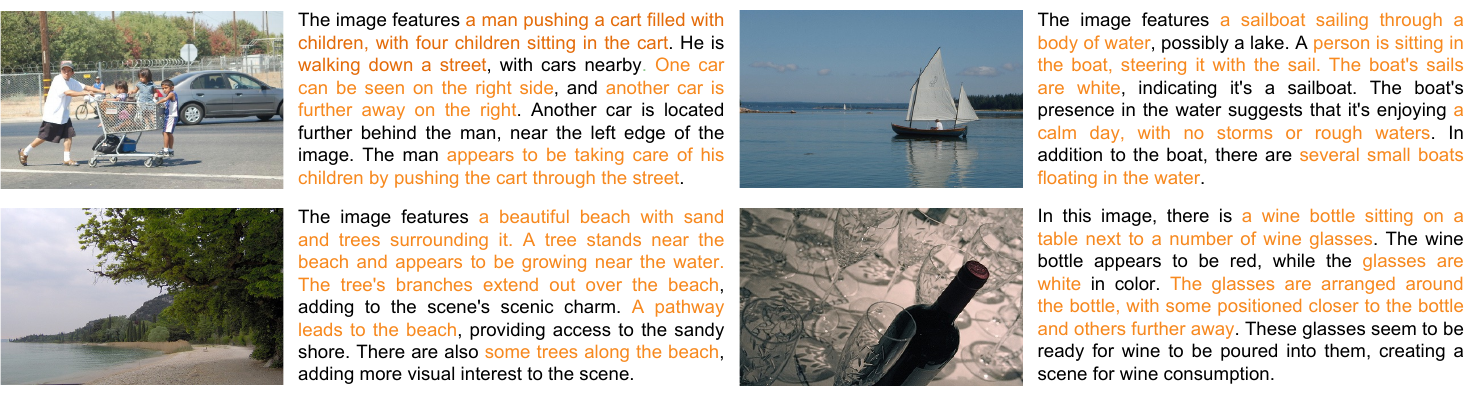} 
  \caption{Image captioning results of LLaVA-1.5-7B compressed by ERC-SVD under 20\% compression ratio. Captions that accurately describe the image content are highlighted in \textcolor{lightergreen}{orange}.}
  \label{fig:image_captioning}
\end{figure}

As shown in Figure~\ref{fig:image_captioning}, LLaVA-1.5-7B compressed with ERC-SVD under 20\% compression ratio is still able to produce image captions that are faithful to the visual content.
This indicates that the model retains its ability to interpret and describe images accurately, despite parameter reduction. 

\begin{figure}[H]
  \centering
  \includegraphics[width=\linewidth]{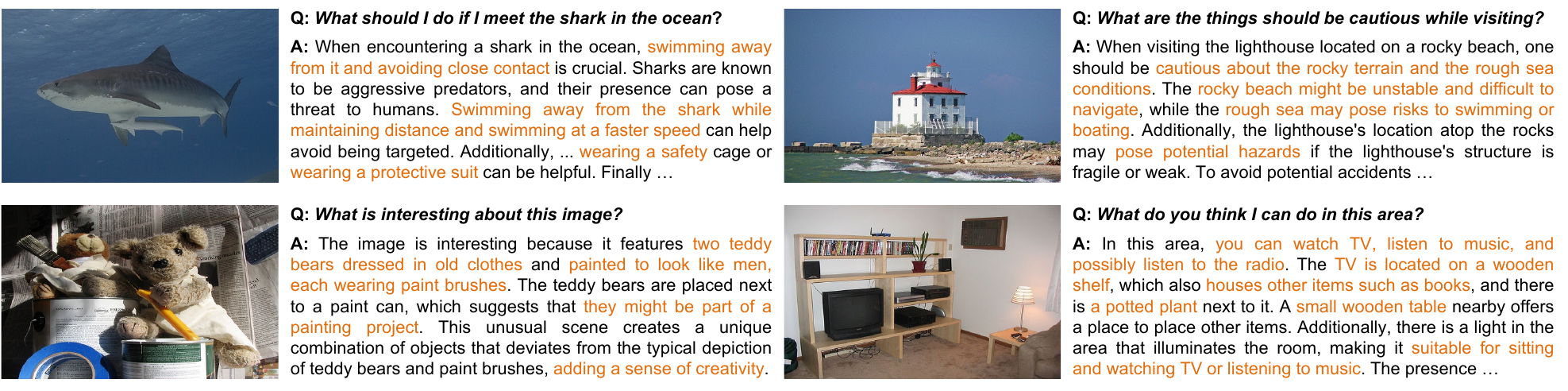} 
  \caption{Visual question answering outputs generated by LLaVA-1.5-7B compressed using ERC-SVD under 20\% compression ratio. Questions (Q) and model answers (A) are provided, correct answers are highlighted in \textcolor{lightergreen}{orange} to emphasize answer quality retention.}
  \label{fig:vqa}
\end{figure}

Figure~\ref{fig:vqa} presents the visual question answering results of LLaVA-1.5-7B compressed by ERC-SVD.
The model retains strong performance across these question types, indicating that its multi-modal reasoning capability remains intact despite the compression.

\subsection{Future Work}
While this work focuses on general LLM compression, the low-rank structures identified via SVD hold significant potential for reasoning LLMs~\cite{jaech2024openai,yang2025qwen3,feng2025efficient}. Beyond autoregressive language tasks, the principles of SVD-based compression are readily transferable to diffusion frameworks~\cite{rombach2022high,zhu2025obs,bai2026dice}. In related tasks, the Transformer-based denoisers often exhibit extreme parameter redundancy. By leveraging ERC-SVD to decompose attention layers, maybe we can accelerate the iterative sampling process, making real-time deployment of generative models on edge devices more feasible without sacrificing structural fidelity. Furthermore, the intersection of model compression and hallucination mitigation~\cite{tonmoy2024comprehensive,zhang2025poison} presents an intriguing research frontier. We hypothesize that low-rank approximations could potentially filter out the noise that contributes to factually incorrect generations, a direction we intend to explore in future work.

\end{document}